\documentclass[fleqn,10pt]{wlscirep}
\usepackage[utf8]{inputenc}
\usepackage{cite}
\usepackage{amsmath,amssymb}
\usepackage{caption}
\usepackage{textcomp}
\usepackage{hyperref}
\usepackage{multicol}
\usepackage{multirow}
\usepackage{interval}
\usepackage{adjustbox}
\usepackage{graphicx}%
\usepackage{multirow}%
\usepackage{amsthm}%
\usepackage{mathrsfs}%
\usepackage[title]{appendix}%
\usepackage{xcolor}%
\usepackage{textcomp}%
\usepackage{manyfoot}%
\usepackage{booktabs}%
\usepackage{algorithm}%
\usepackage{algorithmicx}%
\usepackage{algpseudocode}%
\usepackage{listings}%
\usepackage{array}
\usepackage{ragged2e}
\usepackage{hyperref}
\usepackage{subcaption}
\usepackage{rotating}
\usepackage{pdflscape}
\usepackage{longtable}
\usepackage{colortbl}

\usepackage[T1]{fontenc}
\title{A Hybrid Transformer and Attention Based Recurrent Neural Network for Robust and Interpretable Sentiment Analysis of Tweets}

\author[1,3]{Md Abrar Jahin}
\author[2,3]{Md Sakib Hossain Shovon}
\author[2,3,*]{M. F. Mridha}
\author[4,5,*]{Md Rashedul Islam}
\author[6]{Yutaka Watanobe}
\affil[1]{Department of Industrial Engineering and Management, Khulna University of Engineering \& Technology (KUET), Khulna, 9203, Bangladesh}

\affil[2]{Department of Computer Science, American International University-Bangladesh, Dhaka, 1229, Bangladesh}

\affil[3]{Advanced Machine Intelligence Research (AMIR) Lab, Dhaka, 1229, Bangladesh}

\affil[4]{Offshore AI Development Group, Department of R\&D, Chowagiken Corp., Hokkaido, Japan}

\affil[5]{Department of Computer Science and Engineering, University of Asia Pacific, Dhaka 1216, Bangladesh;}

\affil[6]{Department of Computer Science and Engineering, University of Aizu, Aizu-Wakamatsu 965-8580, Japan}

\affil[*]{Corresponding author: M. F. Mridha (e-mail: firoz.mridha@aiub.edu), Md Rashedul Islam (e-main: rashed.cse@gmail.com).}

\begin{abstract}
Sentiment analysis is a pivotal tool in understanding public opinion, consumer behavior, and social trends, underpinning applications ranging from market research to political analysis. However, existing sentiment analysis models frequently encounter challenges related to linguistic diversity, model generalizability, explainability, and limited availability of labeled datasets. To address these shortcomings, we propose the Transformer and Attention-based Bidirectional LSTM for Sentiment Analysis (TRABSA) model, a novel hybrid sentiment analysis framework that integrates transformer-based architecture, attention mechanism, and recurrent neural network like BiLSTM. The TRABSA model leverages the powerful RoBERTa-based transformer model for initial feature extraction, capturing complex linguistic nuances from a vast corpus of tweets. This is followed by an attention mechanism that highlights the most informative parts of the text, enhancing the model's focus on critical sentiment-bearing elements. Finally, the BiLSTM networks process these refined features, capturing temporal dependencies and improving the overall sentiment classification into positive, neutral, and negative classes. Leveraging the latest RoBERTa-based transformer model trained on a vast corpus of 124M tweets, our research bridges existing gaps in sentiment analysis benchmarks, ensuring state-of-the-art accuracy and relevance. Furthermore, we contribute to data diversity by augmenting existing datasets with 411,885 tweets from 32 English-speaking countries and 7,500 tweets from various US states. This study also compares six word-embedding techniques, identifying the most robust preprocessing and embedding methodologies crucial for accurate sentiment analysis and model performance. We meticulously label tweets into positive, neutral, and negative classes using three distinct lexicon-based approaches and select the best one, ensuring optimal sentiment analysis outcomes and model efficacy. Here, we demonstrate that the TRABSA model outperforms the current seven traditional machine learning models, four stacking models, and four hybrid deep learning models, yielding notable gain in accuracy (94\%) and effectiveness with a macro average precision of 94\%, recall of 93\%, and F1-score of 94\%. Our further evaluation involves two extended and four external datasets, demonstrating the model's consistent superiority, robustness, and generalizability across diverse contexts and datasets. Finally, by conducting a thorough study with SHAP and LIME explainable visualization approaches, we offer insights into the interpretability of the TRABSA model, improving comprehension and confidence in the model's predictions. Our study results make it easier to analyze how citizens respond to resources and events during pandemics since they are integrated into a decision-support system. Applications of this system provide essential assistance for efficient pandemic management, such as resource planning, crowd control, policy formation, vaccination tactics, and quick reaction programs.
\end{abstract}


\begin{document}

\flushbottom
\maketitle
\thispagestyle{empty}

\section*{Introduction}
\label{introduction}
Due to the growth of textual data on social media platforms, news stories, reviews, and consumer feedback, sentiment analysis (SA), a crucial aspect of natural language processing (NLP), has seen growing attention and usage across several domains \cite{sammut_sentiment_2017}. Institutes may get crucial insights into public opinion, consumer preferences, market trends, and brand impression by identifying and analyzing feelings conveyed in text \cite{chaturvedi_sentiment_2017}. Thus, SA is critical in directing marketing initiatives, product development, company strategies, and reputation management \cite{taboada_sentiment_2016}. Additionally, SA is useful in various domains, including politics, healthcare, economics, and the social sciences, where decision-making and policy development depend on a knowledge of human emotions and attitudes.

Despite its broad use, SA still has issues that need more study and creativity. The lack of generalizability and robustness of SA models is one of the primary issues, especially when applying them to various languages, domains, and datasets \cite{kallam_advancements_2023}. Because existing models frequently display different performance levels based on the properties of the data, they are less dependable in real-world situations where the data distribution may change greatly \cite{kumar_spatiotemporal_2022}. Furthermore, SA models' interpretability is still a major worry, particularly in high-stakes scenarios when model predictions are used to make judgments \cite{jawale_interpretable_2020}. Deep learning (DL) models are black-box in nature, which makes it difficult to grasp how these models get to their conclusions. This makes it difficult to implement, impedes trust, and holds people accountable for important decision-making processes.

Inspired by these difficulties, this study aims to develop a strong, broadly applicable, and easily interpreted model of SA that will overcome the shortcomings of current techniques. Through improvements in DL, attention mechanisms, and interpretability methodologies, our goal is to develop a model that performs well on various datasets and offers insights into how it makes decisions. This research advances the area of SA by bridging the gap between model performance, interpretability, and practical application. We aim to improve the trustworthiness, transparency, and usefulness of SA models by employing empirical assessments and interpretability studies. This will enable enterprises to make well-informed decisions by relying on dependable sentiment insights.

Our research aims to fill several critical gaps in the existing literature. First, although SA has received a lot of attention—especially regarding social media data—more robust and interpretable models are still required to classify sentiments across various languages and domains accurately. Many current methods are not transparent, scalable, or generalizable, making it difficult to use them in real-world situations. Furthermore, a notable deficiency in current datasets for SA is the absence of representation for various English language usage patterns. Variations in vocabulary, grammar, and contextual usage of English across national boundaries result in subtle discrepancies in the presentation of distinct emotions. This variability presents a problem for SA models, making it difficult to assess sentiments appropriately in various language circumstances. More advanced methods are required to capture minute semantic subtleties and adjust to changing contextual signals since current models may not comprehend sarcasm, context-dependent sentiment changes, or nuanced sentiment expressions.

This study addresses the need for robust and generalizable SA models by proposing the "Transformer and Attention-based Bidirectional LSTM for Sentiment Analysis (TRABSA)" model. The TRABSA model integrates the strengths of transformer-based architecture and attention mechanisms with recurrent neural networks (RNNs) like bidirectional long short-term memory (BiLSTM) to enhance the performance and adaptability of SA tasks. Our method seeks to capture the broad diversity of English language usage and offer a more thorough knowledge of sentiment expression in various linguistic situations by combining data from several locations into a single dataset. By using this method, TRABSA can more effectively adjust to the subtle differences in English language usage across various groups, which improves the precision and significance of SA findings. We test the performance of the TRABSA model on a range of DL architectures and datasets, including extensive Twitter and external social media datasets, with a particular emphasis on scalability, accuracy, and consistency. Additionally, we do interpretability assessments utilizing the SHAP and LIME approaches to understand the model's decision-making mechanism. We show the TRABSA model's generalizability and robustness through our thorough examination, providing a viable method for SA in various real-world situations.

Our research makes eight-fold key contributions:

\begin{enumerate}
    \item We propose the TRABSA model, a novel hybrid sentiment analysis framework that combines transformer-based architectures, attention mechanisms, and BiLSTM networks to improve sentiment analysis performance.
    \item This research leverages the latest RoBERTa-based transformer model, trained on a vast corpus of 124M tweets, to bridge existing gaps in sentiment analysis benchmarks, ensuring state-of-the-art accuracy and relevance.
    \item We extended the existing dataset by scraping 411,885 tweets from 32 English-speaking countries to include diversity in the Global Twitter COVID-19 Dataset, acknowledging the varied perspectives and discourse across regions. We scraped an additional 7,500 tweets from different states of the USA to deepen geographical representation in the USA Twitter COVID-19 Dataset, allowing for localized insights and analysis.
    \item This article thoroughly compares word embedding techniques, establishing the most robust preprocessing and embedding methodologies essential for accurate sentiment analysis and model performance.
    \item We methodically label tweets using three distinct lexicon-based approaches and rigorously select the most effective one, ensuring optimal sentiment analysis outcomes and model efficacy.
    \item We conduct extensive experiments to assess the TRABSA model's performance on the UK COVID-19 Twitter Dataset, benchmarking against 7 traditional machine learning (ML) models, 4 stacking models, and 4 DL models, demonstrating its superiority and versatility.
    \item We evaluate the TRABSA model's robustness and generalizability across 2 extended and 4 external datasets, showcasing its consistent superiority and applicability across diverse contexts and datasets.
    \item We provide insights into the interpretability of the TRABSA model through rigorous analysis using SHAP and LIME techniques, enhancing understanding and trust in the model's predictions.
\end{enumerate}

The rest of this article is structured as follows: Section "\hyperref[related_work]{Related Works}" reviews state-of-the-art literature in SA using ML-DL and interpretability techniques. In the section "\hyperref[methodology]{Methodology}," the data collection and preprocessing techniques, unsupervised text labeling, implemented ML and DL models for benchmarking, architecture, and methodology of the proposed TRABSA model are described. The section "\hyperref[results]{Results}" presents the experimental setup, evaluation metrics, results, and robustness analysis of the TRABSA model. The SHAP and LIME analysis conducted on the TRABSA model are covered in section "\hyperref[xai]{Interpretability Analysis}". Section "\hyperref[discussions]{Disucssions}" addresses the findings and implications of our investigation. In conclusion, the article is summarized, and future research directions are outlined in section "\hyperref[conclusions]{Conclusions and Future Directions}".

\section*{Related Works}
\label{related_work}
When it comes to classifying data into positive, neutral, and negative sentiment polarity, SA is essential. Exploring a wide range of emotions is the focus of the emerging domains of SA \cite{cambria_affective_2016}. Sentiments can be further classified into categories like satisfaction and rage within certain settings, such as political disputes \cite{dandrea_approaches_2015}. The development of SA approaches with ambivalence management has allowed classifying emotions into distinct classes, including sorrow, anger, anxiety, excitement, and happiness, leading to more nuanced outcomes \cite{wang_multi-level_2020}. While SA has typically focused on textual data, it has expanded to include multimodal SA, which explains data from devices that employ audio- or audio-visual formats \cite{yang_multimodal_2022}. The extension of SA into multimodal analysis highlights its variety and complexity, creating opportunities for a wide range of NLP applications. The variety of options is further highlighted by the fast growth of NLP, fueled by research in neural networks \cite{ray_mixed_2022}. Notably, the development of Neurosymbolic AI, which combines symbolic reasoning and deep learning, offers a viable method of improving NLP capabilities \cite{sarker_neuro-symbolic_2022}, highlighting the various paths NLP research is taking. Lexicon-based methods, ML-based methods, and hybrid techniques are the three main methodologies for solving text categorization and emotion detection challenges. Word polarity is used by lexicon-based approaches, and ML techniques see text analysis as a classification problem that may be further divided into supervised, semi-supervised, and unsupervised learning approaches \cite{chaki_study_2019}. SA results are frequently improved in real-world applications by combining ML with lexicon-based techniques.

In Ahmed \& Ahmed's work, positive and negative emotions were used to classify gathered fake newspapers using a variety of approaches, including TF-IDF, random forest (RF), Naïve Bayes (NB), etc. \cite{ahmed_classification_2023}. According to their results, out of all the classifiers used, the Naïve Bayes classifier had the best accuracy (89.30\%). To identify feelings in the Twitter sentiment 140 datasets, Gaur et al. used TF-IDF feature extraction and the Naïve Bayes Classifier \cite{shakya_twitter_2023}. The model produced improved accuracy (84.44\%) and precision when measured using several performance criteria, such as accuracy, recall, and precision. The COVID-19-related data that Qi \& Shabrina (2023) examined came from Twitter users in major English cities \cite{qi_sentiment_2023}. They conducted a comparative analysis of ML models, including Vader and Textblob, RF, support vector classification (SVC), and multinominal Naïve Bayes (MNB) models. According to the results of their investigation, SVC with TF-IDF demonstrated better accuracy than the other models. To assess opinions about Saudi cruises, Al Sari et al. created three different datasets from social media platforms \cite{al_sari_sentiment_2022}. With oversampled Snapchat data, they used ML techniques, including RF, MLP, NB, voting, SVM, and the n-grams feature extraction approach to reach 100\% accuracy with the RF algorithm. A customized approach for explicit negation detection was presented by Mukherjee et al. \cite{mukherjee_effect_2021}. They used TF-IDF for feature extraction and various ML techniques, including NB, SVM, and Artificial Neural Networks (ANN), to analyze sentiment in Amazon reviews. According to their research, ANNs using negative classifiers had the best accuracy (96.32\%). Using reviews from an international hotel, Noori developed a unique algorithm for classifying client sentiment \cite{noori_classification_2021}. Following the processing of the reviews, document vectors were created using the TF-IDF extractor and trained using SVM, ANN, NB, k-nearest neighbor (KNN), decision tree (DT), and C4.5 models. Outperforming other models, the DT model scored the highest accuracy (98.9\%) with 1800 features. Using N-gram extraction, Zahoor \& Rohilla compared NB, SVM, RF, and long short-term memory networks (LSTM) classifiers on preprocessed datasets \cite{zahoor_twitter_2020}. In most datasets, including the BJP and ML Khattar datasets, NB showed the best accuracy. To turn COVID-19-related tweets into a text corpus and determine the most common terms using N-grams, Samuel et al. used logistic regression (LR) and NB models \cite{samuel_covid-19_2020}. Their results showed that for short tweets, NB and LR had peak accuracy rates of 91\% and 74\%, respectively. For lengthier tweets, both models performed pretty poorly. Using Maximum Entropy (ME), SVM, and LSTM models, Kumar et al. examined the effects of age and gender on customer reviews \cite{kumar_exploring_2020}. LSTM used word2vec, but the NB, ME, and SVM algorithms used Bag of Words (BOW) feature extraction. For female data, the over-50 age group showed the highest accuracy. SVM and MNB with TF-IDF extraction were used by Zarisfi Kermani et al. on four Twitter datasets, and they proposed semantic scoring techniques to represent features in the vector space \cite{zarisfi_kermani_solving_2020}. According to their findings, the suggested technique outperformed the MNB algorithm in three datasets, with the STS dataset showing the greatest MNB performance.

Recent advancements in SA and event detection have introduced several innovative models. DocTopic2Vec, proposed by Truică et al.\cite{truica_topic-based_2021}, enhances document-level SA by combining local and global contexts through document and topic embeddings, outperforming traditional methods. EDSA-Ensemble\cite{petrescu_edsa-ensemble_2024} improves sentiment classification on social media by integrating event detection with SA using an ensemble approach. Petrescu et al.\cite{petrescu_sentiment_2019} bridges network and content analysis by combining event detection with SA, achieving high accuracy in sentiment determination. For imbalanced datasets, Truică and Leordeanu\cite{truica_classification_2017} compare machine learning algorithms, emphasizing the impact of dataset characteristics on classification performance. Lastly, ATESA-BÆRT by Apostol et al.\cite{apostol_atesa-baert_2023} addresses aspect-based SA using a transformers-based ensemble, outperforming existing models in handling reviews with multiple aspects. Additionally, Mitroi et al.\cite{mitroi_sentiment_2020} introduces TOPICDOC2VEC, a new topic-document embedding that combines DOC2VEC and TOPIC2VEC, showing superior performance in polarity detection using game reviews.

In 13 languages with different Indic scripts, Bansal et al. looked at the identification of objectionable language \cite{bansal_transformer_2022}. They assessed four sophisticated transformer-based models and contrasted the Transformer-based method with traditional ML models. Out of all of them, XLM-RoBERTa with BiGRU performed better. Furthermore, adding emoji embeddings to XLM-RoBERTa improved the model's efficacy even further. Due to the combined dataset's code-mixing, training using datasets from 13 Indic languages performed better than training with separate models. Gupta et al. \cite{gupta_emotion_2021} presented a unique emotion analysis approach for real-time COVID-19 tweets, examining eight emotions in different domains. The analysis of tweets from India showed changes in emotional reactions, such as less happiness except for nature. Because of their commitment, teachers' faith in education has grown. In terms of precision and recall, the method by Gupta et al. produced aspect-based graphical and textual summaries from mobile reviews \cite{gupta_aspect-based_2019}, outperforming baseline approaches. Using Twitter data from the Delhi Election 2020, Gupta et al. conducted political echo chamber experiments and investigated the elements that contribute to the creation of echo chambers as well as the role played by users of opposing parties in promoting partisan material \cite{gupta_pendulating_2022}. Gupta and colleagues employed ML algorithms and lexicon-based methods to assess sentiment in Hindi tweets. They found that an integrated CNN-RNN-LSTM model produced an accuracy of 85\% \cite{gupta_toward_2021}. Basiri et al. investigated attitudes about the epidemic in eight different nations using DL classification algorithms, and their findings showed distinct sentiment patterns and relationships with pandemic indicators \cite{basiri_novel_2021}. Using the BERT model, Hayawi et al. achieved excellent accuracy in their ML-based method for spotting COVID-19 vaccination disinformation \cite{hayawi_anti-vax_2022}. Using BERT, Vishwamitra et al. were able to detect hate speech connected to elderly individuals and the Asian community on Twitter during the pandemic \cite{vishwamitra_analyzing_2020}. They were able to identify separate word connections for various hate speech datasets. Before and after the initial COVID-19 case announcement, Chen et al. monitored conversations in Luxembourg about policy and daily life; post-announcement, travel-related issues dominated, perhaps because of the region's large immigrant population \cite{chen_exploratory_2021}. To emphasize changing emotions over time, Kabir et al. used ML for word extraction and emotion categorization in COVID-19 tweets \cite{kabir_emocov_2021}. To categorize COVID-19-related Twitter postings, Valdes et al. created a BERT-based model \cite{valdes_uach-inaoe_2021}, proving the use of domain-specific data for improved performance. During the pandemic, Tziafas et al. used an ensemble architecture to recognize false information, using transformer-based encoders to achieve high accuracy \cite{tziafas_fighting_2021}. Sadia et al. obtained high assessment scores using BERT to conduct SA of COVID-19 tweets \cite{sadia_sentiment_2021}. Song et al. analyzed several facets of misinformation diffusion and created a model to categorize misinformation related to COVID-19 \cite{song_classification_2021}. Using NLP models, Hossain et al. assessed a dataset for COVID-19-related misinformation detection, offering preliminary benchmarks for advancement \cite{hossain_covidlies_2020}. During the COVID-19 lockdown, Chintalapudi et al. used BERT to assess sentiment in Indian tweets, and they showed better accuracy than other models \cite{chintalapudi_sentimental_2021}.

The literature review reveals several gaps in SA research, particularly in the context of COVID-19 and social media sentiment classification. While existing studies have explored SA using various ML algorithms and lexicon-based approaches, comprehensive investigations remain lacking across diverse datasets, including those from different geographic regions and languages. Additionally, previous research has focused on individual datasets or specific domains, neglecting SA models' broader applicability and generalizability. Moreover, studies that directly compare different SA techniques, including DL architectures and ensemble methods, are scarce in identifying the most effective approach across various contexts. The need for interpretability and explainability in SA models is also apparent, with few studies incorporating techniques such as SHAP and LIME for insights into model predictions. These gaps highlight the need for more comprehensive and comparative studies encompassing diverse datasets, languages, and evaluation metrics to advance the SA field effectively.

\section*{Methodology}
\label{methodology}
\begin{figure*}[!ht]
    \centering
    \includegraphics[width=1\linewidth]{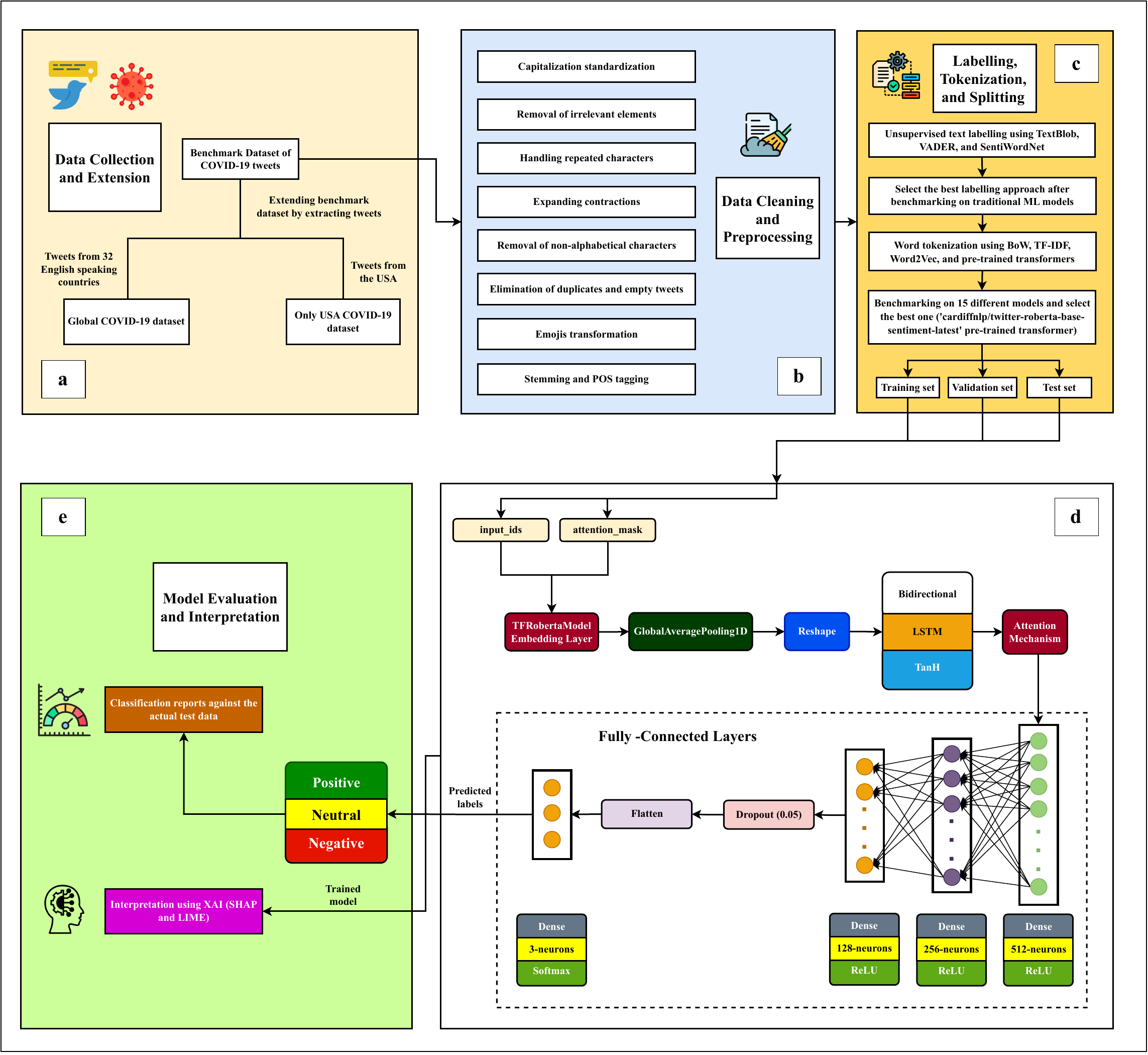}
    \caption{This figure depicts the step-by-step methodological framework proposed for tweet sentiment analysis. It begins with (a) data collection and extension, followed by (b) data cleaning and preprocessing. Subsequently, (c) sentiment labeling into positive, neutral, and negative categories is performed using the `cardiffnlp/twitter-roberta-base-sentiment-latest' pre-trained transformer, and the dataset is split into training, validation, and test sets. The framework proceeds with (d) model development, (e) model benchmarking, and evaluation against baseline models or state-of-the-art approaches. Finally, the process concludes with XAI interpretation techniques applied to gain insights into the model's predictions.}
    \label{fig:trabsa-framework}
\end{figure*}

Our proposed methodological framework outlines a structured approach to SA, encompassing several key stages to ensure robustness and effectiveness in model development and evaluation, as shown in Figure \ref{fig:trabsa-framework}. The first stage involves gathering relevant data for SA. We extend existing datasets by collecting additional tweets from diverse sources to enhance the dataset's representativeness and coverage. Following data collection, we perform cleaning and preprocessing tasks to ensure the quality and consistency of the data. This includes removing noise, expanding contractions, handling duplicates, emojis, and missing tweets, and standardizing text formats. The next step involves labeling the data into positive, neutral, and negative sentiments. We leverage the latest updated RoBERTa-based pre-trained transformer model for tokenization and sentiment labeling, enabling accurate and efficient text data processing. The dataset is divided into training, validation, and test sets after it has been labeled. This enables us to use the test set to assess the model's performance on untested data, refine hyperparameters using the validation set, and train the model on a portion of the data. In this step, we build the SA model based on our suggested hybrid DL architecture. We used Keras Tuner for hyperparameter optimization within the specified search space, focusing on minimizing the validation loss as our objective to achieve the best performance. Following model development, we compare the trained model's performance against baseline models or current state-of-the-art methods. We examine the model's ability to correctly classify sentiments into positive, neutral, and negative categories using a variety of metrics, including accuracy, precision, recall, and F1-score. Finally, we employ XAI techniques to interpret the model's predictions and gain insights into its decision-making process. This involves analyzing the model's internal mechanisms, such as attention weights or feature importance, to understand the factors influencing its predictions and enhance model interpretability.

\subsection*{Data Collection and Preprocessing}
In this section, we provide a comprehensive overview of the data collection and preprocessing procedures undertaken in our study, which laid the foundation for robust SA of tweets.

\subsubsection*{Data Sources} 
This study employed a comprehensive set of seven distinct datasets to facilitate a thorough exploration of SA across various dimensions. These datasets were classified into three main categories: Benchmark, Extended, and External, each serving a unique purpose in our research.

\begin{figure*}[!ht]
  \centering
  \includegraphics[width=1\textwidth, height=0.5\textwidth]{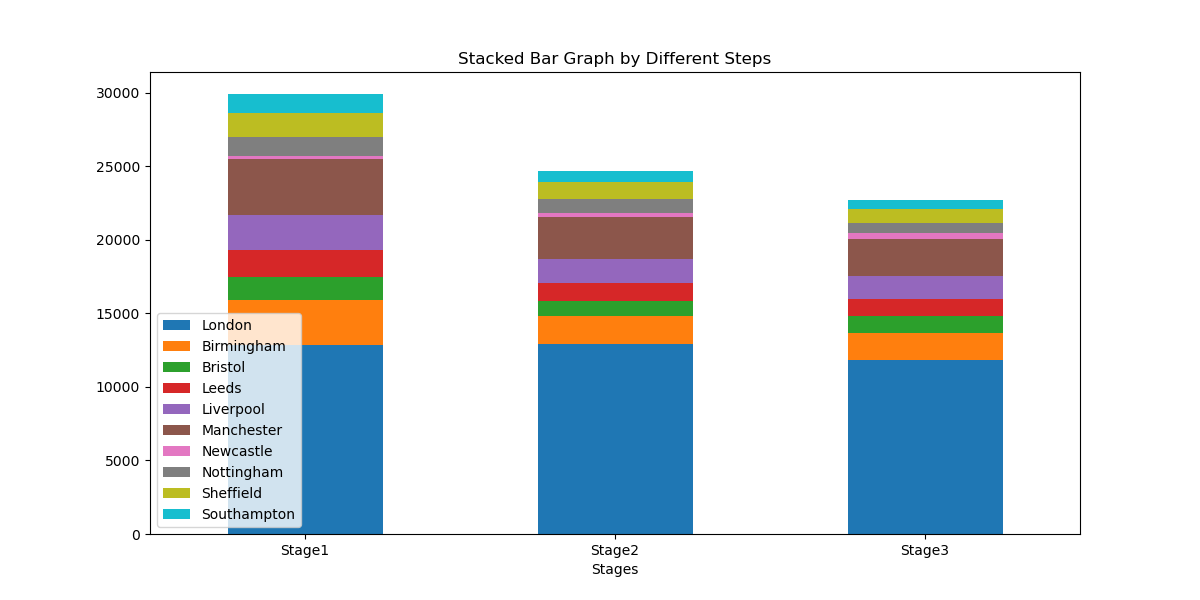}
  \caption{Stacked bar chart showing tweet distribution in three stages of data collection  during third lockdown period from the major cities of the UK.}
  \label{fig4}
\end{figure*}

\textbf{Benchmark Dataset:} The benchmark dataset, which forms the basis of our research, was first assembled and curated by \cite{qi_sentiment_2023}. It functions as a standard by which our proposed model's performance is measured. Interestingly, our model outperformed this benchmark dataset, indicating significant progress in SA. This reference dataset consists of tweets with geotags from well-known cities in the United Kingdom during the third nationwide COVID-19 shutdown. Figure \ref{fig4} shows the tweets gathered from the three stages in the UK. This group of cities includes Greater London, Bristol, South Hampton, Birmingham, Manchester, Liverpool, Newcastle, Leeds, Sheffield, and Nottingham. Over the course of three weeks, from January 6, 2021, to July 18, 2021, 77,332 tweets were gathered. 29,923 tweets were gathered in the first stage, 24,689 in the second, and 22,720 in the third. Major cities such as London, Manchester, Birmingham, and Liverpool were the source of most tweets, with London having the highest count with 37,678. Smaller cities, like Newcastle, had just 852 tweets in a six-month period. The data distribution is in phases, with the first stage having the greatest data and the third stage having the least, as Figure \ref{fig4} illustrates. While Newcastle's contribution was connected with its population and density, London consistently supplied the most data.

\textbf{Extended Datasets:}
To augment our research's cross-cultural dimension and overcome the geographic limitations of the benchmark dataset proposed by Qi and Shabrina, we extended the existing UK COVID-19 Twitter dataset \cite{qi_sentiment_2023}. The tweets were sourced through a combination of data extraction tools, specifically Twint and the Twitter Academic API. These tools were chosen because of their ability to acquire tweets with geolocation information, which is crucial for conducting geographical analyses. However, it should be noted that only a few 1\% of Twitter users actively opt to share their geographic location when composing tweets, and this feature is not enabled by default \cite{sloan_who_2015}. The extended datasets, comprising the Global Twitter COVID-19 Dataset and the USA Twitter COVID-19 Dataset, are publicly available in the \href{https://data.mendeley.com/datasets/2ynwykrfgf/1}{Extended Covid Twitter Datasets} repository \cite{jahin_extended_2023}.

To ensure a comprehensive dataset, we merged the data collected by Twint and the Twitter Academic API. This amalgamation allowed us to access a larger volume of tweets. In identifying tweets related to the COVID-19 pandemic, we employed specific keywords such as "corona" or "covid" in the Twint search configurations and the query field of the Twitter Academic API. This search strategy enabled us to extract tweets and associated hashtags containing these pertinent terms.

\begin{enumerate}
    \item Extended Global COVID-19 Dataset:
This extension involved the comprehensive scraping of 411,885 tweets from 32 English-speaking countries. This dataset expansion allowed us to capture sentiment variations across diverse English-speaking regions, as illustrated in Figure \ref{fig1}. In particular, cities such as the "United States," the "United Kingdom," "Australia," and "New Zealand" exhibit high tweet volumes, while several other cities have comparatively lower tweet counts. Figure \ref{fig2} illustrates word clouds and word frequencies within tweets of the extended datasets. Figure \ref{fig2} (left) represents a visual summary of the most frequently occurring words in a vast dataset related to the COVID-19 pandemic. At the center of this cloud is the word "covid," which dominates with a staggering 226,463 mentions. Other significant terms around it, such as "vaccine," "case," "test," and "people," indicate the key topics and concerns worldwide during the pandemic. Words like "death," "pandemic," and "health" also hold prominence, highlighting the gravity of public health issues. Additionally, terms like "Trump" and "government" suggest the political dimensions entwined with the pandemic discourse. 

\begin{figure*}[!ht]
  \centering
  \includegraphics[width=1\linewidth, height=0.7\linewidth]{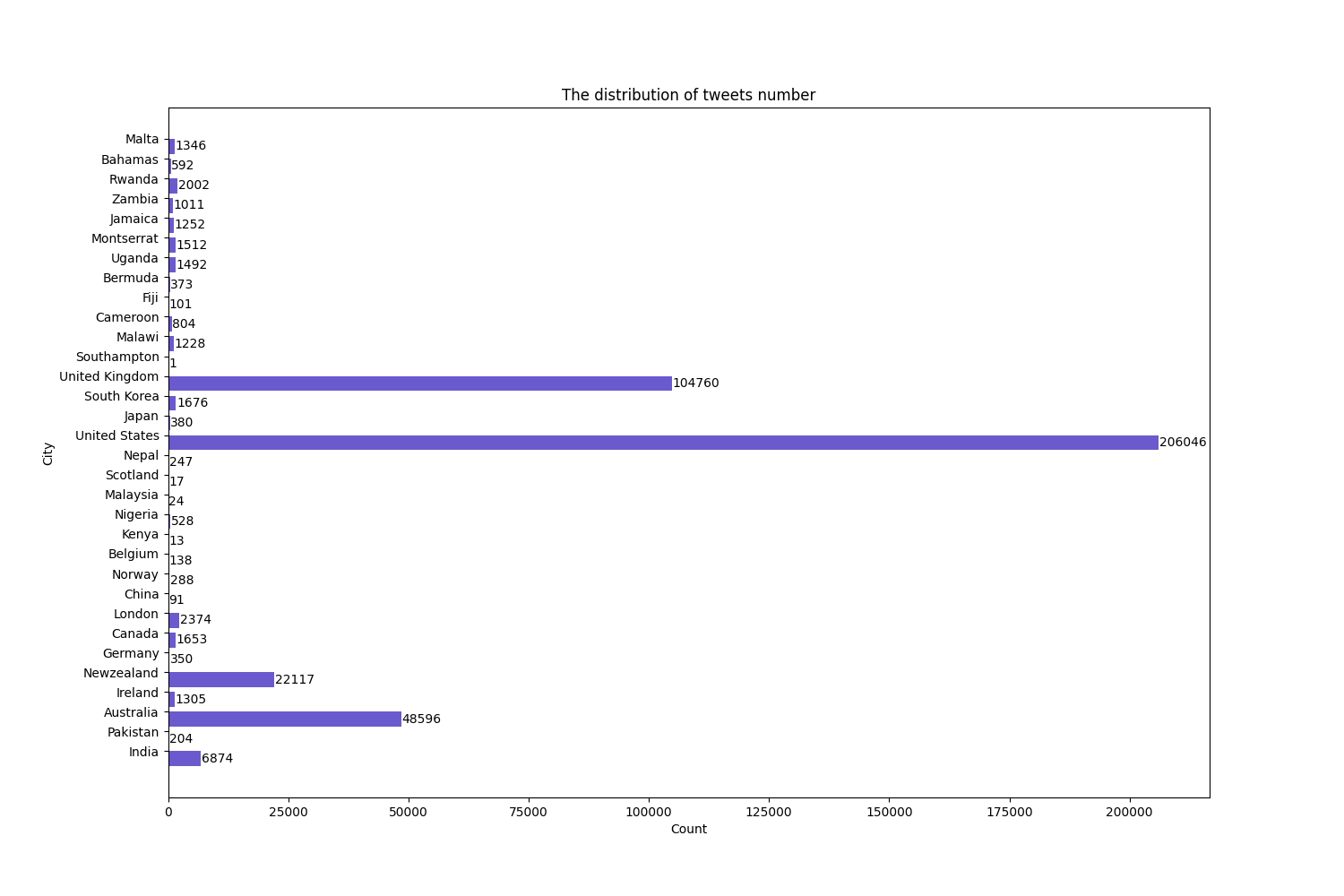}
  \caption{Bar plot showing the distribution of tweets across 32 English-speaking countries.}
  \label{fig1}
\end{figure*}

    \item Extended USA COVID-19 Dataset:
In addition to the international extension, we further enriched our data by creating an extended dataset focusing exclusively on the United States. This dataset comprised 7,500 tweets meticulously scraped from U.S.-based sources. Including this dataset allows for a closer examination of sentiment dynamics within a specific geographical context. Figure \ref{fig2} (right) specific to the United States reveals notable trends and sentiments within the country during the COVID-19 pandemic. In this dataset, words like "corona," "coronavirus," and "covid" are prominent, underlining the ubiquitous presence of these terms in American discussions. Interestingly, words like "fool" and "joke" appear, possibly reflecting a spectrum of attitudes towards the pandemic response. Negative expressions like "shit," "fuck," and "die" also emerge, suggesting the emotional intensity and frustration associated with the situation. Terms like "test" and "case" point towards testing and infection rates concerns, while "april" indicates a temporal reference. 
\end{enumerate}

\begin{figure*}[!ht]
  \centering
  \includegraphics[width=1\textwidth, height=0.45\textwidth]{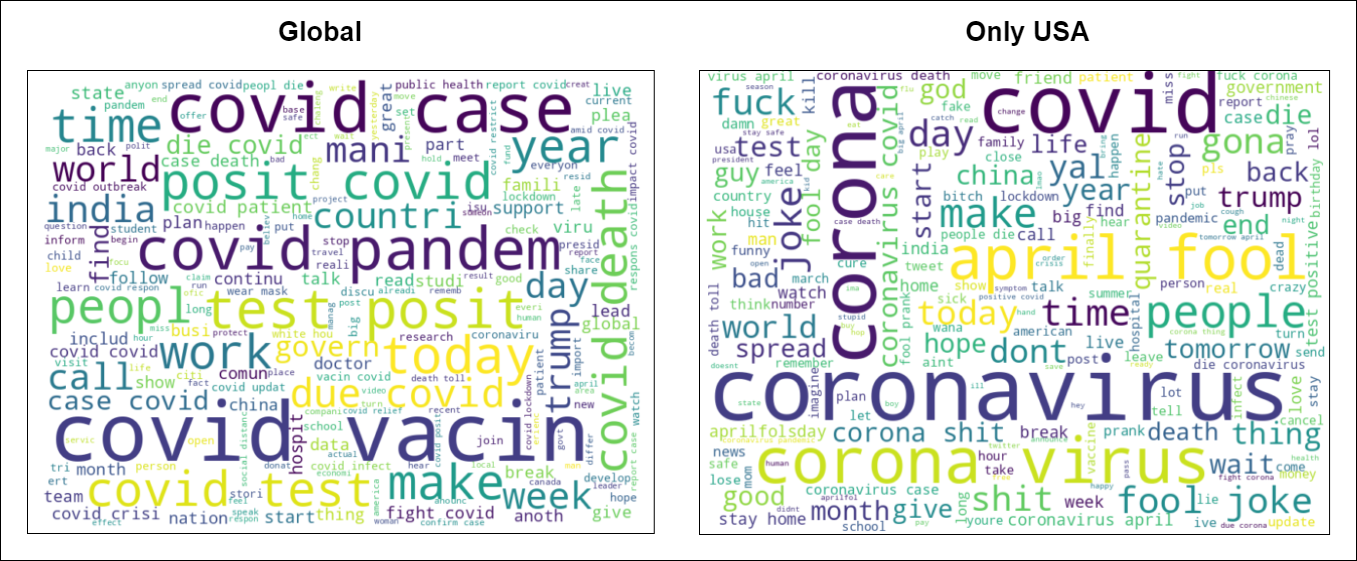}
  \caption{This word cloud visualizes the most frequently occurring words in the "Global" (left) and "Only USA" (right) datasets of COVID-19-related tweets. The size of each word corresponds to its frequency in the dataset.}
  \label{fig2}
\end{figure*}

% \begin{figure*}[!ht]
%   \centering
%   \includegraphics[width=0.95\textwidth, height=0.4\textwidth]{word_frequencies.png}
%   \caption{This line plot displays the word frequency within tweets in the "Global" (left) and "Only USA" (right) COVID-19 datasets. It provides insights into the most commonly used words in the context of COVID-19 discussions among Twitter users.}
%   \label{fig3}
% \end{figure*}

\textbf{External Datasets:}
In order to evaluate the robustness and generalizability of our proposed model, we incorporated four external datasets sourced from Kaggle, including the \href{https://www.kaggle.com/datasets/cosmos98/twitter-and-reddit-sentimental-analysis-dataset}{Twitter and Reddit Dataset}, \href{https://www.kaggle.com/datasets/seriousran/appletwittersentimenttexts}{Apple Dataset}, and \href{https://www.kaggle.com/datasets/crowdflower/twitter-airline-sentiment}{US Airline Dataset} (see \hyperref[data_avail]{Data Availability} section).

\begin{enumerate}
 
    \item Twitter Dataset: This dataset, collected from Twitter, represents diverse tweets covering various topics and subject matter. It enables us to assess the model's adaptability to various Twitter content.

    \item Reddit Dataset: The Reddit dataset encompasses user-generated content from the popular social media platform. Its inclusion allows us to explore sentiment patterns in a different online community, offering valuable insights into the model's versatility.

    \item Apple Dataset: The Apple dataset consists of textual data related to the technology giant Apple Inc. By incorporating this dataset, we aim to analyze sentiment in a specific industry context, providing a more nuanced view of the model's performance.

    \item US Airline Dataset: This dataset is centered around discussions related to U.S. airlines. It allows us to investigate sentiment trends within the context of the aviation industry, adding yet another layer of applicability to our model.
\end{enumerate}

\subsubsection*{Data Cleaning and Preprocessing}
As a previous study has indicated, preprocessing the raw Twitter data was essential to guarantee the accuracy and reliability of our SA because of their informal and unstructured character \cite{naseem_covidsenti_2021}. Our thorough data-cleaning procedure included the following crucial steps:

\begin{enumerate}
    \item Capitalization Standardization: To prevent the recognition of identical words with varying capitalization as distinct, we uniformly converted all text to lowercase. This step was crucial for consistent word recognition.

    \item Removal of Irrelevant Elements: We methodically eliminated any superfluous content that has no bearing on SA, including hashtags (\#subject), stated usernames (@username), and any hyperlinks beginning with "www," "http," or "https." We also removed terms that were less than two characters and stop words. Stop words, though common in text, often lack significant sentiment polarity. Despite being often used in texts, stop words frequently lack strong emotive polarity. It's important to note that negations like "not" and "no" were kept in because removing them may change the sense of whole sentences.

    \item Handling Repeated Characters: Some users utilize repeating characters in their tweets to highlight intense feelings. Words not present in standard lexicons were transformed into their correct forms to standardize such expressions. For instance, "sooooo goooood" was normalized to "so good."

    \item Extending Contractions: Removing punctuation after a contraction, such "isn't" or "don't," presented difficulties. They were expanded into their full forms to maintain the meaningfulness of contractions. For instance, "isn’t" became "is not."

    \item Elimination of Non-Alphabetical Characters: All punctuation, numerals, and special symbols were removed, along with all other non-alphabetical characters and symbols. These extraneous characters had the potential to interfere with feature extraction.

    \item Elimination of Duplicates and Empty Tweets: We identified and removed duplicated or empty tweets to ensure data integrity, creating a clean and consistent dataset.
    \item Emojis Transformation: Given the prevalence of emojis in tweets to express sentiment and emotion, we adopted the `demojize()' function from Python's emoji module to transform emojis into their corresponding textual meanings. This enhancement was especially beneficial for improving the accuracy of SA.

    \item Advanced Cleaning for Specific Approaches: Depending on the SA approach employed, additional cleaning steps, such as stemming and Part-of-Speech (POS) tagging, were applied. These steps were particularly relevant for methods relying on resources like SentiWordNet.
   
\end{enumerate}

By rigorously implementing these data-cleaning procedures, we ensured that our SA was conducted on a high-quality dataset, minimizing noise and optimizing the extraction of meaningful sentiment features.

\subsubsection*{Word Embeddings} 
We used various word embedding approaches in this work to extract contextual and semantic information from our textual material. Our NLP tasks performed much better thanks to these embeddings. The word embedding techniques we used in our studies are summarized in the next subsections.

\textbf{Bag-of-Words (BoW):} BoW is a classic technique for word representation. It transforms tweets into vectors by counting the frequency of words in each tweet. While it doesn't capture word order or context, it provides a straightforward and interpretable way to represent text data. To utilize the BoW approach, we employed the `CountVectorizer` function from the scikit-learn library.

\textbf{Term Frequency-Inverse Document Frequency (TF-IDF):} We utilized the TF-IDF embeddings by employing the `TfidfVectorizer` function from the scikit-learn module, which allocates weights to words according to their significance in individual tweets and their scarcity throughout the complete dataset. Additionally, it made it easier to down-weight frequent keywords, which allowed our models to concentrate on more informative words.

\textbf{Word2Vec:} One of Word2Vec's advantages is that it can record semantic similarities between words, which makes text data analysis more sophisticated. Using neural networks, it represents words as dense vectors in a continuous vector space. The `word\_tokenize` function from the NLTK library was utilized to tokenize our tweets, as it allows for the breakdown of sentences into individual words. Using the Gensim package, a Word2Vec model was produced with the vector size, window size, and skip-gram model set. By using a continuous vector space, this approach was able to express words as vectors. Two methods were used to encode full tweets as vectors: sum vectorization and average vectorization.

\textbf{Pre-trained Transformers:} In our research, we harnessed the power of pre-trained transformer-based models from the Hugging Face Transformers library to leverage contextual embeddings for text data. Three distinct transformer models were employed, each bringing unique capabilities to the analysis. The `distilbert-base-uncased' model, known for its efficiency and lightweight nature, was selected for its suitability in scenarios where computational resources are constrained. It produces context-aware word embeddings that consider each word's left and right context. We used a state-of-the-art `cardiffnlp/twitter-roberta-base-sentiment-latest' model, updated in 2022, to capture sentiment-specific nuances in a tweet. This model was trained on an extensive dataset of approximately 124 million tweets collected from January 2018 to December 2021. This model was designed for English text and was a robust foundation for our SA endeavors. Additionally, we incorporated `sentence-transformers/all-MiniLM-L6-v2,' a sophisticated tool that transforms tweets into a dense vector space of 384 dimensions. It transforms entire sentences into fixed-dimensional vectors while maintaining semantic information.

Although the code implementation for each transformer followed a similar structure, the choice of model brought diversity to our experimentation, enabling us to explore the impact of contextual embeddings on our text classification task. To tokenize and process our text data effectively, we employed the model's associated tokenizer, incorporating techniques such as padding and truncation to ensure consistent input lengths. The tokenized data was then efficiently processed on the GPU for optimal computational performance. We further harnessed the model's capabilities to extract the hidden states associated with the `[CLS]` token, which often encapsulates the comprehensive context of the text.

\subsection*{Unsupervised Text Labeling}
In ML, labeling vast amounts of text data manually can be time-consuming. To address this challenge and expedite the labeling process, we leveraged lexicon-based methods, specifically TextBlob, VADER, and SentiWordNet, to automatically assign sentiment scores to tweets. Our sentiment classification scheme employed three categories: positive (assigned a value of 1), negative (assigned -1), and neutral (assigned 0).

We used the BoW method implemented with the CountVectorizer from the scikit-learn library to convert text data from our benchmark dataset into a matrix of word frequencies. We conducted a comprehensive evaluation to determine the effectiveness of our unsupervised labeling approach. The performance of seven traditional base ML models was evaluated against sentiment scores derived from each of the three lexicon approaches: TextBlob, VADER, and SentiWordNet. Our evaluation unveiled that TextBlob consistently outperformed VADER and SentiWordNet regarding accuracy across all implemented ML models, as shown in Table \ref{tab1}. Hence, we used TextBlob-based labels for further benchmarking.

\begin{table*}[!ht]
\centering
\caption{Performance evaluation of unsupervised sentiment labeling approaches}
\label{tab1}
\begin{tabular}{|l|c|c|c|} 
\hline
\multirow{2}{*}{\textbf{ML Model}} & \multicolumn{3}{c|}{\textbf{Accuracy of Lexicon based methods}}  \\ 
\cline{2-4}
                                    & TextBlob      & VADER & SentiWordNet                             \\ 
\hline
RF                                  & \textbf{67\%} & 65\%  & 57\%                                     \\ 
\hline
NB                                  & \textbf{66\%} & 59\%  & 61\%                                     \\ 
\hline
SVM                                 & \textbf{73\%} & 66\%  & 72\%                                     \\ 
\hline
GBM                                 & \textbf{79\%} & 71\%  & 69\%                                     \\ 
\hline
LGBM                                & \textbf{79\%} & 73\%  & 70\%                                     \\ 
\hline
XGBoost                             & \textbf{76\%} & 70\%  & 71\%                                     \\ 
\hline
Catboost                            & \textbf{74\%} & 69\%  & 68\%                                     \\
\hline
\end{tabular}
\end{table*}

\subsection*{Models Used}
\subsubsection*{Traditional ML Models} 
Our analysis encompasses a diverse set of models traditional ML models, including traditional base models, their stacked ensembles, and voting classifiers. The aim was to comprehensively evaluate the performance of these models using different text representations. The following models were used:

\begin{enumerate}
    \item Random Forest (RF): RF is an ensemble learning method that aggregates the predictions of multiple decision trees. It is known for its robustness and ability to handle high-dimensional data.

    \item Naive Bayes (NB): NB is a probabilistic classifier based on Bayes' theorem, assuming independence among features. It is particularly well-suited for text classification tasks.

    \item Support Vector Machine (SVM): SVM is a powerful classifier that aims to find a hyperplane that best separates data points in a high-dimensional space. It is effective for both linear and non-linear classification.

    \item Gradient Boosting Machine (GBM): GBM is an ensemble learning technique that builds decision trees sequentially, focusing on the mistakes of the previous trees. It often leads to strong predictive performance.

    \item LightGBM (LGBM): LightGBM is a gradient-boosting framework for efficiency and speed. It uses a histogram-based approach for tree construction.

    \item XGBoost: XGBoost is another popular gradient-boosting library known for its scalability and performance optimization. It has been widely used in various ML competitions.

    \item CatBoost: CatBoost is a gradient-boosting library specializing in categorical feature support. It is known for its ability to tackle categorical data effectively.

    \item LGBM + K-Nearest Neighbors (KNN) + Multi-Layer Perceptron (MLP): We explored an ensemble approach by combining LGBM with KNN and MLP to leverage the strengths of different algorithms.

    \item RF + KNN + MLP: Similarly to the previous ensemble, we combined RF with KNN and MLP to diversify our modeling approach further.

    \item GBM + RF Stacking Classifier: Stacking is an ensemble technique where multiple models' predictions are combined using another model. Here, we stack GBM and RF to improve predictive accuracy potentially.

    \item GBM + RF Voting Classifier: Voting classifiers combine the predictions of multiple models by majority voting. We used this ensemble technique to take advantage of the collective wisdom of GBM and RF.
\end{enumerate}

We explored various combinations of word embeddings and text representations for each model, including BoW, TF-IDF, Word2Vec, and pre-trained transformer models for text tokenization. These different representations allowed us to assess the impact of text preprocessing on model performance and gain insights into which models were most effective for sentiment classification.

\subsubsection*{Deep Neural Networks (DNNs)} 
To rigorously evaluate tweet sentiment classification, we employed a diverse set of DNN models, each with distinct architectural characteristics. We utilized the Keras Tuner for hyperparameter tuning across these models to ensure optimal performance. The search space included LSTM layer units ranging from 128 to 768, dense layer units from 64 to 512, dropout rates from 0.1 to 0.5, and learning rates from $1\times10^{-5}$ to $1\times10^{-2}$. Using Keras Tuner's Random Search method, we identified the best parameters for each model, significantly enhancing their performance. This systematic exploration, coupled with comprehensive text representations and optimized hyperparameters, provided valuable insights into the performance of various DNN architectures.

Our initial model, the `Single-Dense Layered Neural Network,' started with a transformer for feature extraction. A single dense layer with 512 units and ReLU activation captured high-level representations. The simplicity of this architecture allowed us to establish a baseline for performance comparison.

Building upon this foundation, we introduced the `3 dense layers of neural network. After global averaging of the transformer's outputs, three sequential dense layers were introduced, with decreasing units (512, 256, 128) to refine feature representations progressively. In particular, dropout regularization was applied after the first dense layer, enhancing model robustness.

We introduced the `BiLSTM + 3 Hidden Dense Layers' model to explore the nuances of SA texts further. This architecture incorporated a BiLSTM layer, which is a special type of RNN, enabling the network to capture sequential dependencies in the input data. Following the BiLSTM layer, three additional dense layers (512, 256, and 128 units) were used to distill the features further. Dropout was applied to enhance model generalization.

Lastly, we explored hybrid architecture with the `BiLSTM + CNN' model. Here, the model combined the strengths of a BiLSTM layer with convolutional layers. The convolutional layers, with 64 filters and varying kernel sizes, added a spatial perspective to feature extraction. Subsequently, two dense layers (128 and 64 units) were introduced to further process the extracted features.

\subsubsection*{Proposed TRABSA Model}
\begin{figure*}[!ht]
    \centering
    \includegraphics[width=1\linewidth]{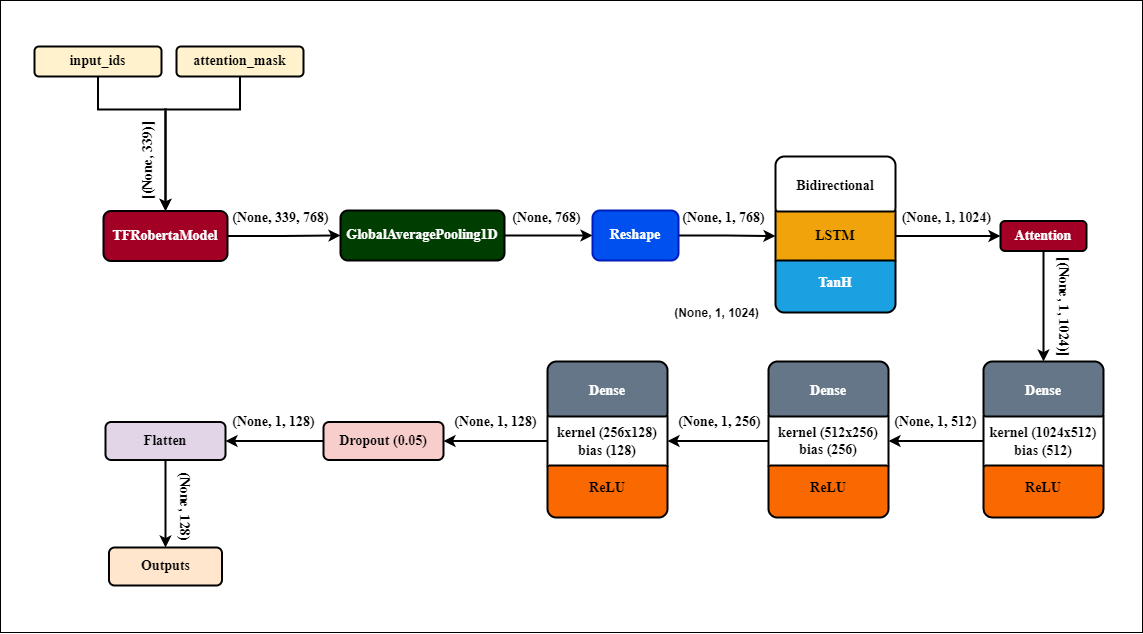}
    \caption{Model architecture of the proposed TRABSA model.}
    \label{fig:trabsa-archi}
\end{figure*}

The proposed TRABSA model presents a systematic and effective architecture for SA, as shown in Figure \ref{fig:trabsa-archi}. The model utilizes the `cardiffnlp/twitter-roberta-base-sentiment-latest' pre-trained transformer, capitalizing on its contextual understanding of the text. The architecture begins with input layers, including `input\_ids' and `attention\_mask,' where a maximum sequence length of 256 tokens is utilized. RoBERTa is used for its excellent performance in tasks involving natural language understanding. It encodes the input tweet text and generates contextual embeddings.

The input data is then organized into a TensorFlow dataset, shuffled, and batched for training. The batch size is set to 16 to facilitate efficient training. The labels are one-hot encoded to prepare them for multiclass classification. The dataset is split into training and validation sets for rigorous evaluation. In optimizing the TRABSA model, Keras Tuner plays a crucial role by systematically exploring a well-defined search space to fine-tune various hyperparameters. The search space includes the number of units in the BiLSTM layer, ranging from 128 to 512, which balances model complexity and efficiency. Additionally, three Dense layers are tuned with units ranging from 128 to 768, 64 to 512, and 32 to 256, respectively, affecting the model’s capacity and computational demands. The dropout rate varies from 0.1 to 0.5 to prevent overfitting, while the learning rate is explored logarithmically between 1e-5 and 1e-3 to optimize convergence speed. The tuning process employs Random Search to sample various hyperparameter combinations, providing an efficient way to explore the space without exhaustive search. After running multiple trials, the tuner identifies the best hyperparameter settings (see Figure \ref{fig:trabsa-archi}) based on validation loss, ensuring an optimized balance between performance and computational efficiency.

The core architecture of the TRABSA model is based on the pre-trained RoBERTa-base model architecture with specific enhancements:

\begin{itemize}
 
    \item \textit{Input Layers:} Two input layers are defined - `input\_ids' and `attention\_mask,' which receive the tokenized input tweet sequences and their corresponding attention masks, ensuring proper handling of padded fixed-length sequences.

    \item \textit{Transformer Embeddings:} The transformer produces contextual embeddings, which capture rich information about the text. These embeddings are then subjected to `Global Average Pooling' to reduce the dimensionality while retaining essential features. A reshaping operation is applied to prepare the data for subsequent layers.

    \item \textit{BiLSTM:} The BiLSTM layer is an advanced type of RNN designed to enhance sequence modeling by capturing contextual information from both directions in a sequence. In this model, the BiLSTM layer is configured with 512 units in each direction, totaling $512\times2$ units. The forward LSTM network processes the sequence from start to end, while the backward LSTM network processes it from end to start. This bidirectional approach allows the BiLSTM layer to integrate information from both past and future tokens, providing a more nuanced understanding of the text.

    \item \textit{Self-Attention Mechanism:} An attention layer is incorporated, which applies self-attention to the output of the BiLSTM. This mechanism allows the model to weigh the importance of different parts of the input sequence, which can be critical for understanding the nuances of sentiment in tweets.

    \item \textit{Dense Layers:} A series of densely connected layers are added to capture complex patterns and relationships within the data. While there are multiple dense layers, their architecture plays a crucial role. A 512-unit dense layer with ReLU activation serves as the primary feature extractor, followed by two more dense layers (256 and 128 units) to refine representations progressively. A dropout layer with a 0.05 dropout rate contributes to regularization and helps prevent overfitting.

    \item \textit{Flatten Layer:} After processing through the dense layers, the output tensor is flattened to a 1D vector.

    \item \textit{Classifier Head:} The classifier head consists of a dense layer with three units, using the softmax activation function. It produces the input tweet's final sentiment classification probabilities (positive, negative, or neutral).
    
\end{itemize}

The model is compiled using the Adam optimizer with a learning rate of $4\times10^{-5}$ and categorical cross-entropy loss. Categorical accuracy is used as the evaluation metric. We included model checkpointing, early stopping, and callbacks to optimize model training. The early stopping mechanism monitors validation loss and restores the best weights to prevent overfitting. The training process involves fitting the model on the training dataset and validating it on the validation dataset for 50 epochs; however, due to the early stopping mechanism, the iterations stop after 23 epochs.

\begin{figure*}[!ht]
    \centering
    \includegraphics[width=1\linewidth]{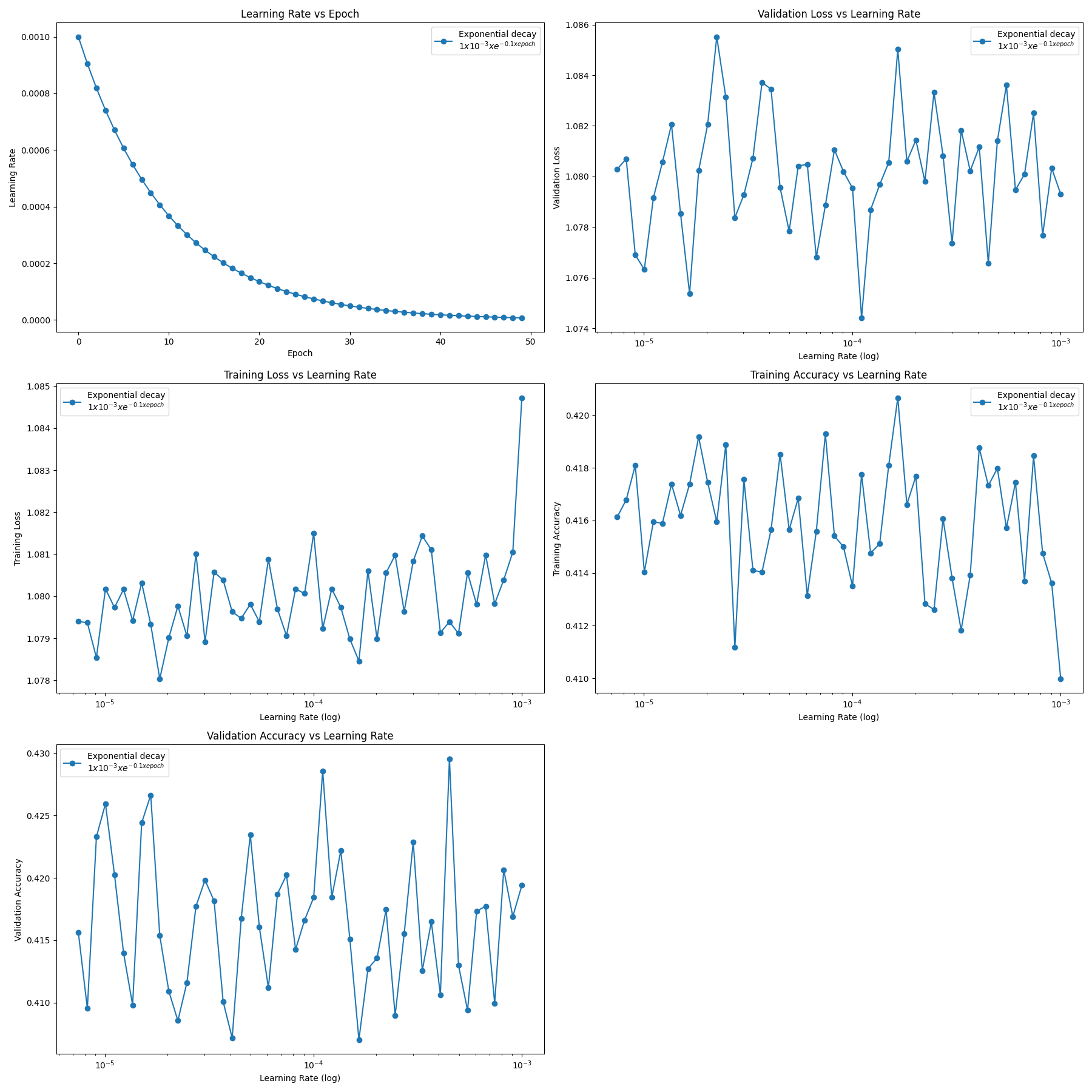}
    \caption{This figure depicts the learning rate scheduler callback function utilized during model training. It visualizes the decayed learning rate as the epoch increases, alongside the corresponding training and validation accuracy, as well as training and validation loss, plotted against variable learning rates.}
    \label{fig:lr}
\end{figure*}

\begin{figure*}[!ht]
    \centering
    \includegraphics[width=1\linewidth]{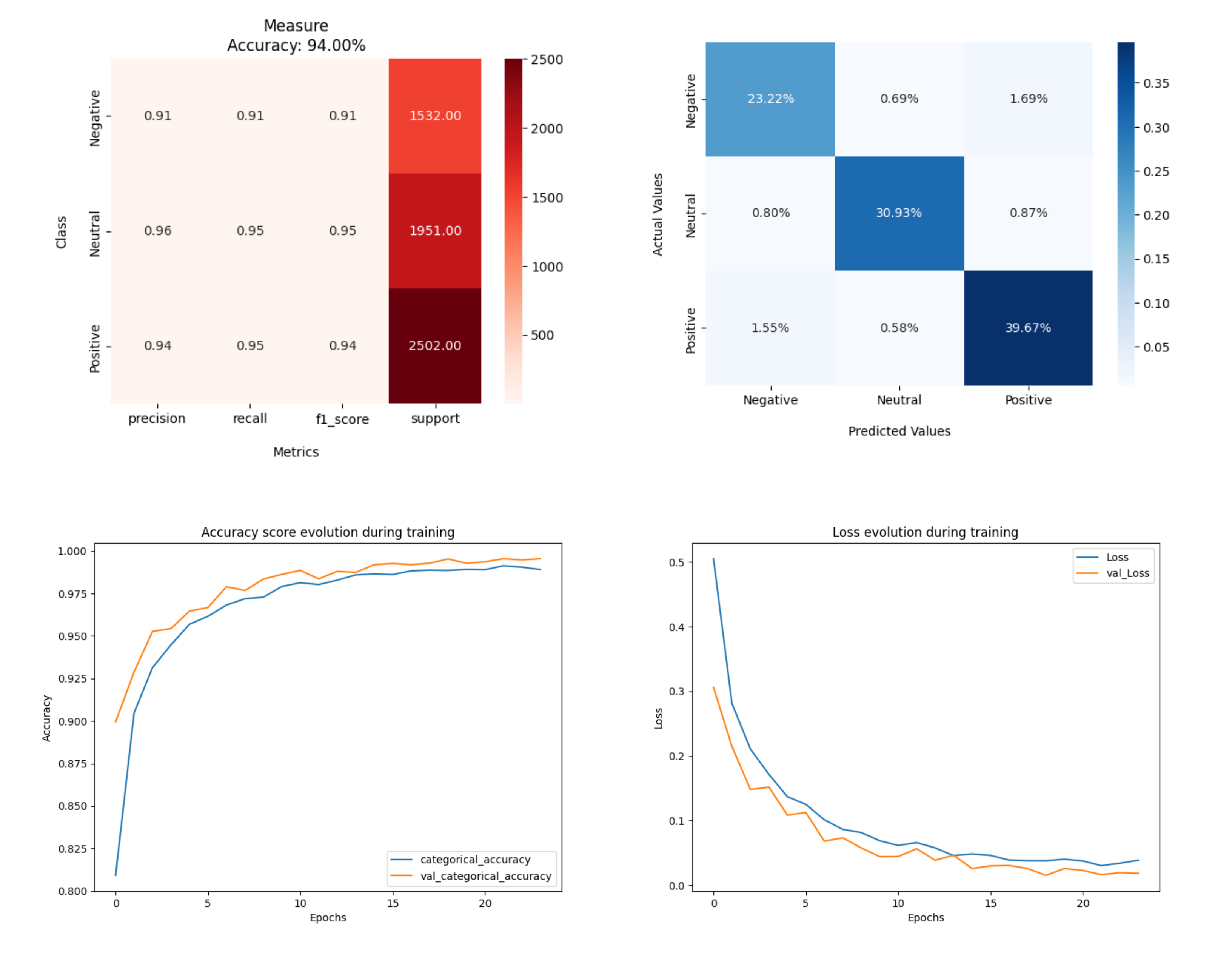}
    \caption{Masterplot illustrating the TRABSA model's performance starting from the top right and proceeding clockwise including the (a) Classification Metrics Plot, (b) Confusion Matrix, (c) Loss vs Epoch Curve, (d) Training and Validation Accuracy vs Epoch Curve.}
    \label{fig:loss-confusion}
\end{figure*}

The learning rate scheduler callback function adjusted the learning rate during our model training. The function calculates the learning rate for each epoch based on an initial learning rate and an exponential decay factor, which controls the rate at which the learning rate decreases over epochs. In this specific implementation, the exponential decay formula is utilized, where the learning rate is multiplied by the exponent of a negative constant $k=0.1$ multiplied by the epoch number. As the epoch increases, the learning rate exponentially decreases, allowing for a gradual reduction in the learning rate during training. This technique helps optimize the training process by fine-tuning the learning rate to improve model convergence and performance over successive epochs, as shown in Figure \ref{fig:lr}.

Figure \ref{fig:loss-confusion} comprehensively evaluates the TRABSA model's performance in tweet SA. The top left corner showcases the model's classification metrics plot, illustrating precision, recall, and F1-score metrics for each sentiment class. Moving clockwise, the confusion matrix provides a detailed analysis of the model's classification performance by comparing predicted sentiment labels with actual labels. The loss vs. epoch curve illustrates its training and validation loss over successive epochs, while the training and validation accuracy vs. epoch curve depicts the model's learning progress and convergence. Together, these visualizations offer insights into the TRABSA model's classification accuracy, convergence, and optimization process, aiding researchers in assessing its performance and identifying areas for improvement.

The TRABSA model's architecture combines the strengths of RoBERTa's contextual embeddings, BiLSTM's sequence modeling, attention mechanisms for capturing interdependencies, and a well-designed set of dense layers to improve SA accuracy for tweet data. This architecture demonstrated superior performance compared to other models, making it a noteworthy addition to the field of SA.

\section*{Results}
\label{results}
\subsection*{Experimental Setup}
Table \ref{tab2} overviews the hardware and software specifications used in our ML and DL experiments. It includes details on the CPU, GPU, TPU, RAM, Python version, and essential libraries utilized to conduct the research.

\begin{table*}[!ht]
\centering
\caption{Details of hardware and software specifications}
\label{tab2}
\resizebox{\linewidth}{!}{%
\begin{tabular}{|>{\hspace{0pt}}m{0.3\linewidth}|>{\hspace{0pt}}m{0.788\linewidth}|} \hline
\textbf{Resources Used} & \textbf{Specifications} \\ \hline
Intel(R) Xeon(R) CPU & x86 with a clock frequency of 2 GHz, 4 vCPU cores, 18GB \\ \hline
NVIDIA T4 x2 GPU & 2560 Cuda cores, 16 GB \\ \hline
Google TPU & 8 TPU v3 cores. 128 GB \\ \hline
RAM & 16 GB DDR4 \\ \hline
Python & version 3.10.12 \\ \hline
Libraries & numpy, pandas, matplotlib, seaborn, nltk, TensorFlow, keras, PyTorch, genism, scikit-learn, joblib, transformers, re, string, shap, scipy, lime \\ \hline
\end{tabular}
}
\end{table*}

\subsection*{Evaluation Metrics}
We use a variety of assessment indicators in our research to determine our models' overall success. These measurements show how well the models correctly categorize sentiments across various classifications. The F1-score, accuracy, precision, recall, and macro-average metrics are among the crucial assessment measures employed. These are computed using the following definitions of true positives (TP), false positives (FP), and false negatives (FN):

\begin{equation}
    \text{Precision} = \frac{TP}{TP + FP}
\end{equation}
\begin{equation}
    \text{Precision} = \frac{TP}{TP + FP}
\end{equation}
\begin{equation}
    \text{F1-score} = 2 \times \frac{\text{Precision} \times \text{Recall}}{\text{Precision} + \text{Recall}}
\end{equation}

Recall, also known as sensitivity, quantifies the percentage of properly identified positive occurrences among all real positive instances, whereas precision describes the proportion of correctly classified positive instances among all cases projected as positive. The F1-score, which is derived from the harmonic mean of accuracy and recall, offers a fair indicator of model performance.

The accuracy metric quantifies the percentage of properly identified occurrences out of all instances, as follows:

\begin{equation}
\text{Accuracy} = \frac{TP + TN}{TP + TN + FP + FN}
\end{equation}

Where TN represents true negatives.

Furthermore, we calculate the F1-score, recall, and macro-average accuracy to present a comprehensive evaluation of the model's performance across all sentiment classes:

\begin{equation}
\text{Macro Average Precision} = \frac{1}{N} \sum_{i=1}^{N} \text{Precision}_i
\end{equation}
\begin{equation}
\text{Macro Average Recall} = \frac{1}{N} \sum_{i=1}^{N} \text{Recall}_i
\end{equation}
\begin{equation}
\text{Macro Average F1-score} = \frac{1}{N} \sum_{i=1}^{N} \text{F1-score}_i
\end{equation}

Where \(N\) represents the total number of sentiment classes.

Together, these evaluation measures offer a thorough analysis of the model's performance employed in this research, assisting in the choice and improvement of SA models.

\subsection*{Results}
Analyzing the performance of various models across different word embedding techniques reveals significant variations in their effectiveness for SA tasks. Among the traditional models, GBM and LightGBM consistently demonstrate strong performance across all word embedding techniques, achieving macro average F1-scores ranging from 79\% to 83\%. Pre-trained transformer models like BERT and RoBERTa exhibit competitive performance, with the RoBERTa model showcasing high accuracy across various configurations, especially when combined with advanced neural network architectures. Notably, the TRABSA model outperforms all other models across all evaluation metrics, demonstrating exceptional macro average precision (94\%), macro average recall (93\%), macro average F1-score (94\%), and accuracy of 94\% (see Table \ref{tab3}). This significant improvement underscores the efficacy of the TRABSA framework in SA tasks, surpassing even state-of-the-art transformer models like RoBERTa.

We used BoW, TFIDF, word2vec, BERT, SBERT, and RoBERTa as word embeddings with several ML and two DL models. When we noticed a significant improvement in the accuracy (above 80\%) while using RoBERTa word embeddings with simple DL architectures, we decided to implement more complex hybrid DL models: BiLSTM+3 Hidden Layers NN, BiLSTM+CNN, and TRABSA.

%\onecolumn
\footnotesize
\begin{longtable}[!ht]{|c|l|p{1.5cm}|p{1.4cm}|p{1.3cm}|p{1.2cm}| p{1cm} | p{1cm} |}

\caption{Comprehensive 10-fold cross-validated mean performance evaluation metrics with standard deviations for various models and embeddings, including time performance (training and inference times)}
\label{tab3}\\
\hline
\textbf{Word Embedding} & \textbf{Model} & \textbf{Macro Average Precision} & \textbf{Macro Average Recall} & \textbf{Macro Average F1-score} & \textbf{Accuracy} & \textbf{Training Time (s)} & \textbf{Inference Time (s)}\\
\hline
\endfirsthead

\caption*{\textbf{Table \ref{tab3} (Continued):} Comprehensive 10-fold cross-validated mean performance evaluation metrics with standard deviations for various models and embeddings, including time performance (training and inference times)}\\
\hline
\textbf{Word Embedding} & \textbf{Model} & \textbf{Macro Average Precision} & \textbf{Macro Average Recall} & \textbf{Macro Average F1-score} & \textbf{Accuracy} & \textbf{Training Time (s)} & \textbf{Inference Time (s)}\\
\hline
\endhead

BoW & RF & 75\%$\pm$1\% & 61\%$\pm$2\% & 61\%$\pm$2\% & 67\%$\pm$1\% &  6.73 & 0.03  \\ 
\hline
BoW & NB & 63\%$\pm$1\% & 62\%$\pm$2\% & 62\%$\pm$2\% & 64\%$\pm$1\% & 0.82 & 0.01 \\ 
\hline
BoW & SVM & 76\%$\pm$2\% & 76\%$\pm$2\% & 76\%$\pm$2\% & 77\%$\pm$1\% & 4.35 & 0.30 \\ 
\hline
BoW & GBM & 82\%$\pm$1\% & 79\%$\pm$1\% & 79\%$\pm$1\% & 81\%$\pm$1\% & 262.81 & 0.02 \\ 
\hline
BoW & LGBM & 83\%$\pm$1\% & 81\%$\pm$1\% & 81\%$\pm$1\% & 83\%$\pm$1\% & 35.54 & 0.01  \\ 
\hline
BoW & XGBoost & 77\%$\pm$0\% & 75\%$\pm$1\% & 75\%$\pm$1\% & 78\%$\pm$0\% & 86.50 & 0.02 \\ 
\hline
BoW & Catboost & 80\%$\pm$0\% & 74\%$\pm$1\% & 75\%$\pm$1\% & 77\%$\pm$1\% & 536.16 & 0.02 \\ 
\hline
BoW & LGBM+KNN+MLP & 82\%$\pm$1\% & 81\%$\pm$2\% & 81\%$\pm$2\% & 84\%$\pm$0\% & 1422.53 & 45.43 \\ 
\hline
BoW & RF+KNN+MLP & 74\%$\pm$1\% & 73\%$\pm$2\% & 74\%$\pm$1\% & 75\%$\pm$1\% & 1212.43 & 40.54 \\ 
\hline
BoW & GBM+RF Stacking Classifier & 78\%$\pm$0\% & 76\%$\pm$1\% & 76\%$\pm$1\% & 78\%$\pm$0\% & 1364.06 & 50.57 \\ 
\hline
BoW & GBM+RF Voting Classifier & 78\%$\pm$0\% & 69\%$\pm$1\% & 70\%$\pm$1\% & 72\%$\pm$1\% & 1250.37 & 48.30 \\ 
\hline
BoW & Single Hidden Layer NN & 71\%$\pm$4\% & 69\%$\pm$4\% & 69\%$\pm$4\% & 72\%$\pm$4\% & 646.34 & 24.53 \\ 
\hline
BoW & 3 Hidden Layers NN & 68\%$\pm$4\% & 67\%$\pm$4\% & 67\%$\pm$4\% & 70\%$\pm$4\% & 904.64 & 25.35 \\ 
\hline
TFIDF & RF & 69\%$\pm$1\% & 60\%$\pm$3\% & 60\%$\pm$3\% & 66\%$\pm$2\% & 6.75 & 0.04 \\ 
\hline
TFIDF & NB & 65\%$\pm$1\% & 56\%$\pm$3\% & 56\%$\pm$3\% & 61\%$\pm$2\% & 0.89 & 0.02 \\ 
\hline
TFIDF & SVM & 74\%$\pm$2\% & 70\%$\pm$3\% & 70\%$\pm$3\% & 73\%$\pm$2\% & 5.15 & 0.46 \\ 
\hline
TFIDF & GBM & 80\%$\pm$1\% & 77\%$\pm$3\% & 77\%$\pm$3\% & 79\%$\pm$1\% & 273.51 & 0.05 \\ 
\hline
TFIDF & LGBM & 80\%$\pm$0\% & 77\%$\pm$3\% & 78\%$\pm$3\% & 80\%$\pm$1\% & 36.12 & 0.02 \\ 
\hline
TFIDF & XGBoost & 68\%$\pm$1\% & 67\%$\pm$1\% & 67\%$\pm$1\% & 70\%$\pm$1\% & 87.22 & 0.03 \\ 
\hline
TFIDF & Catboost & 74\%$\pm$1\% & 72\%$\pm$1\% & 72\%$\pm$1\% & 74\%$\pm$1\% & 542.44 & 0.03 \\ 
\hline
TFIDF & LGBM+KNN+MLP & 79\%$\pm$0\% & 77\%$\pm$1\% & 77\%$\pm$1\% & 79\%$\pm$0\% & 1532.37 & 46.24 \\ 
\hline
TFIDF & RF Bagging & 76\%$\pm$0\% & 48\%$\pm$1\% & 43\%$\pm$1\% & 56\%$\pm$1\% & 764.34 & 35.34 \\ 
\hline
TFIDF & RF+KNN+MLP & 75\%$\pm$1\% & 71\%$\pm$3\% & 72\%$\pm$2\% & 74\%$\pm$0\% & 1254.65 & 42.75 \\ 
\hline
TFIDF & GBM+RF Stacking Classifier & 77\%$\pm$0\% & 76\%$\pm$1\% & 76\%$\pm$1\% & 78\%$\pm$0\% & 1352.53 & 52.65 \\ 
\hline
TFIDF & GBM+RF Voting Classifier & 75\%$\pm$0\% & 69\%$\pm$2\% & 70\%$\pm$1\% & 73\%$\pm$0\% & 1283.23 & 50.75\\ 
\hline
TFIDF & Single Hidden Layer NN & 68\%$\pm$3\% & 67\%$\pm$4\% & 67\%$\pm$4\% & 69\%$\pm$2\% & 650.34 & 22.43  \\ 
\hline
TFIDF & 3 Hidden Layers NN & 93\%$\pm$3\% & 92\%$\pm$4\% & 92\%$\pm$3\% & 93\%$\pm$4\% & 954.64 & 27.53 \\ 
\hline
word2vec & RF & 51\%$\pm$1\% & 49\%$\pm$2\% & 49\%$\pm$2\% & 56\%$\pm$1\% & 6.82 & 0.05 \\ 
\hline
word2vec & NB & 41\%$\pm$1\% & 34\%$\pm$1\% & 19\%$\pm$1\% & 36\%$\pm$1\% & 1.03 & 0.04\\ 
\hline
word2vec & SVM & 67\%$\pm$0\% & 49\%$\pm$2\% & 45\%$\pm$3\% & 57\%$\pm$1\% & 5.78 & 0.57 \\ 
\hline
word2vec & GBM & 51\%$\pm$1\% & 50\%$\pm$2\% & 50\%$\pm$2\% & 56\%$\pm$1\% & 282.10 & 0.08 \\ 
\hline
word2vec & LGBM & 53\%$\pm$1\% & 49\%$\pm$1\% & 47\%$\pm$1\% & 57\%$\pm$1\% & 37.41 & 0.03 \\ 
\hline
word2vec & XGBoost & 55\%$\pm$1\% & 50\%$\pm$2\% & 48\%$\pm$2\% & 56\%$\pm$1\% & 88.60 & 0.03 \\ 
\hline
word2vec & Catboost & 71\%$\pm$0\% & 49\%$\pm$1\% & 45\%$\pm$1\% & 57\%$\pm$1\% & 553.37 & 0.04 \\ 
\hline
word2vec & LGBM+KNN+MLP & 53\%$\pm$1\% & 46\%$\pm$2\% & 41\%$\pm$3\% & 53\%$\pm$1\% & 1448.76 & 47.34 \\ 
\hline
word2vec & RF+KNN+MLP & 55\%$\pm$1\% & 49\%$\pm$3\% & 43\%$\pm$3\% & 57\%$\pm$1\% & 1345.53 & 55.23 \\ 
\hline
word2vec & GBM+RF Stacking Classifier & 46\%$\pm$2\% & 45\%$\pm$2\% & 45\%$\pm$2\% & 51\%$\pm$1\% & 1412.64 & 53.73 \\ 
\hline
word2vec & GBM+RF Voting Classifier & 47\%$\pm$1\% & 47\%$\pm$1\% & 47\%$\pm$1\% & 53\%$\pm$1\% & 1350.23 &  28.78 \\ 
\hline
word2vec & Single Hidden Layer NN & 36\%$\pm$5\% & 45\%$\pm$4\% & 40\%$\pm$4\% & 53\%$\pm$3\% & 704.65 & 28.34 \\ 
\hline
word2vec & 3 Hidden Layers NN & 38\%$\pm$4\% & 48\%$\pm$4\% & 42\%$\pm$4\% & 56\%$\pm$3\% & 1034.89 & 33.38 \\ 
\hline
BERT & RF & 53\%$\pm$2\% & 51\%$\pm$3\% & 49\%$\pm$3\% & 58\%$\pm$1\% & 805.43 & 45.53 \\ 
\hline
BERT & NB & 50\%$\pm$3\% & 51\%$\pm$2\% & 50\%$\pm$2\% & 53\%$\pm$1\% & 1.12 & 0.07 \\ 
\hline
BERT & SVM & 52\%$\pm$2\% & 52\%$\pm$2\% & 52\%$\pm$2\% & 56\%$\pm$1\% &  6.12 & 0.76 \\ 
\hline
BERT & GBM & 56\%$\pm$1\% & 54\%$\pm$2\% & 54\%$\pm$2\% & 60\%$\pm$2\% & 291.11 & 0.11 \\ 
\hline
BERT & LGBM & 58\%$\pm$2\% & 57\%$\pm$2\% & 57\%$\pm$2\% & 61\%$\pm$2\% & 37.66 & 0.03 \\ 
\hline
BERT & XGBoost & 56\%$\pm$2\% & 56\%$\pm$2\% & 56\%$\pm$2\% & 60\%$\pm$2\% & 89.10 & 0.04 \\ 
\hline
BERT & Catboost & 56\%$\pm$2\% & 49\%$\pm$2\% & 46\%$\pm$2\% & 57\%$\pm$2\% & 562.29 & 0.05 \\ 
\hline
BERT & LGBM+KNN+MLP & 61\%$\pm$1\% & 59\%$\pm$2\% & 59\%$\pm$2\% & 62\%$\pm$1\% & 1623.65 & 65.34 \\ 
\hline
BERT & RF+KNN+MLP & 60\%$\pm$1\% & 58\%$\pm$2\% & 57\%$\pm$2\% & 62\%$\pm$1\% & 1443.22 & 60.44 \\ 
\hline
BERT & GBM+RF Stacking Classifier & 55\%$\pm$2\% & 54\%$\pm$2\% & 54\%$\pm$2\% & 56\%$\pm$1\% & 1523.75 & 70.23 \\ 
\hline
BERT & GBM+RF Voting Classifier & 60\%$\pm$1\% & 58\%$\pm$2\% & 58\%$\pm$2\% & 66\%$\pm$0\% & 1452.45 & 67.85 \\ 
\hline
BERT & Single Hidden Layer NN & 61\%$\pm$3\% & 58\%$\pm$4\% & 59\%$\pm$4\% & 62\%$\pm$2\% & 945.53 & 35.64 \\ 
\hline
BERT & 3 Hidden Layers NN & 62\%$\pm$4\% & 60\%$\pm$4\% & 60\%$\pm$4\% & 64\%$\pm$4\% & 1305.39 & 45.49 \\ 
\hline
SBERT & RF & 61\%$\pm$0\% & 48\%$\pm$3\% & 44\%$\pm$4\% & 54\%$\pm$2\% & 6.95 & 0.06 \\ 
\hline
SBERT & NB & 53\%$\pm$2\% & 40\%$\pm$3\% & 32\%$\pm$4\% & 44\%$\pm$2\% & 1.14 & 0.09 \\ 
\hline
SBERT & SVM & 56\%$\pm$1\% & 57\%$\pm$2\% & 56\%$\pm$2\% & 58\%$\pm$1\% & 6.28 & 0.81 \\ 
\hline
SBERT & GBM & 55\%$\pm$2\% & 51\%$\pm$2\% & 49\%$\pm$2\% & 55\%$\pm$2\% & 302.43 & 0.15 \\ 
\hline
SBERT & LGBM & 55\%$\pm$2\% & 52\%$\pm$2\% & 50\%$\pm$2\% & 56\%$\pm$2\% & 38.18 & 0.04 \\ 
\hline
SBERT & XGBoost & 56\%$\pm$2\% & 56\%$\pm$2\% & 56\%$\pm$2\% & 57\%$\pm$2\% & 90.53 & 0.06 \\ 
\hline
SBERT & Catboost & 69\%$\pm$0\% & 48\%$\pm$3\% & 42\%$\pm$4\% & 53\%$\pm$2\% & 577.54 & 0.05 \\ 
\hline
SBERT & LGBM+KNN+MLP & 55\%$\pm$2\% & 49\%$\pm$2\% & 47\%$\pm$2\% & 53\%$\pm$2\% & 1734.23 & 68.64 \\ 
\hline
SBERT & RF+KNN+MLP & 56\%$\pm$1\% & 54\%$\pm$2\% & 54\%$\pm$2\% & 57\%$\pm$1\% & 1522.42 & 63.43 \\ 
\hline
SBERT & GBM+RF Stacking Classifier & 44\%$\pm$2\% & 44\%$\pm$2\% & 43\%$\pm$2\% & 48\%$\pm$1\% & 1623.86 & 72.57 \\ 
\hline
SBERT & GBM+RF Voting Classifier & 53\%$\pm$2\% & 50\%$\pm$2\% & 50\%$\pm$2\% & 54\%$\pm$2\% & 1553.54 & 74.67 \\ 
\hline
SBERT & Single Hidden Layer NN & 55\%$\pm$3\% & 55\%$\pm$4\% & 54\%$\pm$3\% & 59\%$\pm$3\% & 954.64 & 38.48 \\ 
\hline
SBERT & 3 Hidden Layers NN & 56\%$\pm$4\% & 57\%$\pm$3\% & 57\%$\pm$3\% & 59\%$\pm$2\% & 1402.54 & 48.48 \\ 
\hline
RoBERTa & RF & 62\%$\pm$2\% & 62\%$\pm$2\% & 62\%$\pm$2\% & 64\%$\pm$1\% & 7.23 & 0.08 \\ 
\hline
RoBERTa & NB & 63\%$\pm$1\% & 55\%$\pm$2\% & 52\%$\pm$3\% & 54\%$\pm$2\% & 1.16 & 1.05 \\ 
\hline
RoBERTa & SVM & 65\%$\pm$1\% & 65\%$\pm$1\% & 65\%$\pm$1\% & 67\%$\pm$0\% & 6.76 & 0.88 \\ 
\hline
RoBERTa & GBM & 63\%$\pm$2\% & 62\%$\pm$3\% & 63\%$\pm$2\% & 65\%$\pm$2\% & 314.44 & 0.17 \\ 
\hline
RoBERTa & LGBM & 64\%$\pm$2\% & 63\%$\pm$3\% & 64\%$\pm$2\% & 66\%$\pm$1\% & 38.89 & 0.05 \\ 
\hline
RoBERTa & XGBoost & 65\%$\pm$1\% & 63\%$\pm$2\% & 63\%$\pm$2\% & 66\%$\pm$1\% & 91.14 & 0.07 \\ 
\hline
RoBERTa & Catboost & 63\%$\pm$2\% & 62\%$\pm$3\% & 61\%$\pm$4\% & 64\%$\pm$2\% & 583.28 & 0.07 \\ 
\hline
RoBERTa & LGBM+KNN+MLP & 66\%$\pm$2\% & 66\%$\pm$2\% & 65\%$\pm$3\% & 66\%$\pm$2\% & 1823.93 & 72.23 \\ 
\hline
RoBERTa & RF+KNN+MLP & 58\%$\pm$3\% & 55\%$\pm$2\% & 55\%$\pm$2\% & 60\%$\pm$1\% & 1654.54 & 67.76 \\ 
\hline
RoBERTa & GBM+RF Stacking Classifier & 61\%$\pm$3\% & 61\%$\pm$3\% & 61\%$\pm$3\% & 63\%$\pm$2\% & 1705.36 & 75.96 \\ 
\hline
RoBERTa & GBM+RF Voting Classifier & 60\%$\pm$2\% & 60\%$\pm$2\% & 59\%$\pm$3\% & 60\%$\pm$2\% & 1653.78 & 73.47  \\ 
\hline
RoBERTa & Single Hidden Layer NN & 84\%$\pm$4\% & 84\%$\pm$4\% & 84\%$\pm$4\% & 84\%$\pm$4\% & 1349.46 & 154.39 \\ 
\hline
RoBERTa & 3 Hidden Layers NN & 84\%$\pm$3\% & 83\%$\pm$4\% & 83\%$\pm$4\% & 84\%$\pm$3\% & 1898.05 & 153.64 \\ 
\hline
RoBERTa & BiLSTM+3 Hidden Layers NN & 84\%$\pm$3\% & 84\%$\pm$3\% & 84\%$\pm$3\% & 85\%$\pm$2\% & 3404.54 &  148.43 \\ 
\hline
RoBERTa & BilSTM+CNN & 83\%$\pm$2\% & 81\%$\pm$4\% & 82\%$\pm$3\% & 83\%$\pm$2\% & 5328.73 & 178.64 \\ 
\hline
\textbf{RoBERTa} & \textbf{Proposed TRABSA Model} & \textbf{94\%$\pm$1\%} & \textbf{93\%$\pm$2\%} & \textbf{94\%$\pm$1\%} & \textbf{94\%$\pm$1\%} & 3675.21 &  147.14 \\
\hline

\end{longtable}

\begin{minipage}{\textwidth}
\renewcommand{\thempfootnote}{\arabic{mpfootnote}}
\footnotesize
\begin{longtable}[!ht]{|c|l|p{1.5cm}|p{1cm}|p{1.1cm}|p{1cm}| p{1cm} | p{1cm} |}

\caption{Ablation test of the proposed TRABSA model}
\label{tab:ablation}\\
\hline
\textbf{Word Embedding} & \textbf{Model} & \textbf{Macro Average Precision} & \textbf{Macro Average Recall} & \textbf{Macro Average F1-score} & \textbf{Accuracy} & \textbf{Training Time (s)} & \textbf{Inference Time (s)}\\
\hline

RoBERTa & w/o\footnote{"w/o" stands for "without," indicating the absence of the specific component in the TRABSA model configuration.} BiLSTM + Attention + 3 Dense Layers & 76\%$\pm$2\% & 75\%$\pm$3\% & 75\%$\pm$2\% & 76\%$\pm$3\% & 1290.53 & 115.68 \\ %direct zero shot
\hline 
RoBERTa & w/o BiLSTM + Attention Layer & 81\%$\pm$1\% & 80\%$\pm$1\% & 80\%$\pm$1\% & 80\%$\pm$2\% & 3500.23 & 169.34 \\ %3 dense
\hline 
RoBERTa & w/o Attention Layer & 85\%$\pm$2\% & 84\%$\pm$3\% & 84\%$\pm$2\% & 84\%$\pm$2\% & 2250.15 & 145.85 \\ %bilstm+3dense
\hline
RoBERTa & w/o 3 Dense Layers & 82\%$\pm$2\% & 81\%$\pm$3\% & 81\%$\pm$3\% & 82\%$\pm$3\% & 2245.64 & 150.14 \\ %bilstm+attention
\hline
RoBERTa & w/o Dropout Layer & 84\%$\pm$1\% & 84\%$\pm$1\% & 84\%$\pm$1\% & 84\%$\pm$2\% & 2200.70 & 140.45 \\ %whole model without dropout
\hline 
RoBERTa & LSTM instead of BiLSTM Layer & 83\%$\pm$2\% & 82\%$\pm$3\% & 82\%$\pm$2\% & 83\%$\pm$3\% & 2380.90 & 142.30 \\ %whole model without bidirectional wrapper
\hline
\end{longtable}
\end{minipage}

\normalsize
Compared to the best-performing traditional models like GBM (81\%), LightGBM (83\%), stacked LGBM+KNN+MLP (84\%), and advanced hybrid state-of-the-art BiLSTM+3 Hidden Layers NN (85\%), the TRABSA model achieves an impressive accuracy of 94\%, indicating a substantial improvement of at least 9\% over the closest competitors. The TRABSA model consistently outperforms others in accuracy and macro average precision, recall, and F1-score, demonstrating its robustness and effectiveness across different evaluation criteria. Even when compared to sophisticated neural network architectures like Single Hidden Layer NN (84\%) and 3 Hidden Layers NN (84\%), the TRABSA model exhibits a remarkable performance boost of $\approx10\%$, further emphasizing its superiority in SA tasks. The TRABSA model's outstanding performance underscores its hybrid architecture's efficacy, which integrates transformer-based mechanisms, attention mechanisms, and BiLSTM networks to capture nuanced sentiment patterns effectively. The findings suggest that while pre-trained transformer models like BERT and RoBERTa offer competitive performance, customized architectures like TRABSA tailored specifically for SA tasks can yield substantial accuracy and predictive power improvements.

The ablation study presented in Table \ref{tab:ablation} offers a detailed analysis of the proposed TRABSA model by examining the effects of removing or altering key components. The full model, not shown in this table, achieves the highest performance, with macro average precision and F1-score reaching 94\%, macro average recall at 93\%, and accuracy at 94\%. This highlights the effectiveness of the BiLSTM, Attention, and Dense layers in capturing complex patterns from the data.

When the BiLSTM, Attention, and Dense layers are all removed, the model's performance drastically drops to 76\% for precision, 75\% for recall, 75\% for F1-score, and 76\% for accuracy. The reduction in training time ($\approx$1290 seconds) and inference time ($\approx$116 seconds) indicates the computational savings from eliminating these layers, but the performance decline reveals their critical role in generalizing well to the data.

In the model where only the Attention layer is removed, the performance sees a significant improvement compared to the full removal of BiLSTM and Dense layers. This configuration achieves an 85\% macro average precision, 84\% recall, and 84\% F1-score, with an accuracy of 84\%. The training time is $\approx$2250 seconds, and the inference time is $\approx$146 seconds, suggesting that while the Attention layer adds value, its absence doesn’t drastically impair the model’s ability to capture temporal dependencies, likely due to the strength of the BiLSTM.

Interestingly, replacing the BiLSTM with a unidirectional LSTM results in a slight drop in performance across all metrics, with precision at 83\%, recall at 82\%, F1-score at 82\%, and accuracy at 83\%. As expected, the training and inference times ($\approx$2381 seconds and $\approx$142 seconds, respectively) show that unidirectional LSTMs are slightly more efficient but at the cost of losing the bidirectional context offered by the BiLSTM.

Moreover, the model without the Dropout layer exhibits similar performance to the attention-removed configuration, maintaining 84\% across all metrics, but with a slightly faster training time ($\approx$2201 seconds) and inference time ($\approx$140 seconds). This suggests that Dropout contributes to regularization, but its absence does not significantly degrade performance, possibly due to the robustness of the RoBERTa embeddings and other layers.

The study underscores the importance of BiLSTM and Attention layers for optimal performance while also demonstrating the computational costs associated with these enhancements. The model without these components, while more computationally efficient, sacrifices accuracy, confirming the balance between complexity and performance in the proposed TRABSA model.

%\twocolumn

\subsection*{Robustness Test of TRABSA Model}
The TRABSA model consistently demonstrates robustness and generalizability across datasets and DL architectures, as evidenced by its consistent performance metrics. Across various datasets, including the Global COVID-19 Dataset, the USA COVID-19 Dataset, the External Twitter Dataset, the Reddit Dataset, the Apple Dataset, and the US Airline Dataset, the TRABSA model consistently achieves high macro average precision, recall, F1-score, and accuracy values. For instance, in the Global COVID-19 Dataset, the TRABSA model attains an impressive 98\% macro average precision, recall, F1-score, and accuracy. Similarly, the USA COVID-19 dataset maintains high scores, obtaining 87\% in terms of accuracy. The trend continues across External datasets, with the TRABSA model consistently performing exceptionally well, achieving an average accuracy of 97\% on the Twitter Dataset, 95\% on the Reddit Dataset, 90\% on the Apple Dataset, and 96\% on the US Airline Dataset (see Table \ref{tab:robustness}). These consistent and high-performance metrics underscore the reliability and effectiveness of the TRABSA model across diverse datasets and DL architectures, reaffirming its robustness and generalizability in SA tasks.

\begin{table}[!ht]
\caption{Generalizability and robustness of the proposed TRABSA model on both the extended and external datasets}
\label{tab:robustness}
\resizebox{\columnwidth}{!}{%
\begin{tabular}{|l|l|l|cccc|c|c|}
\hline
\multicolumn{1}{|c|}{\multirow{2}{*}{\textbf{Dataset Type}}} &
  \multicolumn{1}{c|}{\multirow{2}{*}{\textbf{Dataset Name}}} &
  \multicolumn{1}{c|}{\multirow{2}{*}{\textbf{DL Models}}} &
  \multicolumn{4}{c|}{\textbf{Evaluation Metrics}} &
  \multirow{2}{*}{\textbf{Training Time (s)}} &
  \multirow{2}{*}{\textbf{Inference Time (s)}} \\ \cline{4-7}
\multicolumn{1}{|c|}{} &
  \multicolumn{1}{c|}{} &
  \multicolumn{1}{c|}{} &
  \multicolumn{1}{c|}{\textbf{Macro Average Precision}} &
  \multicolumn{1}{c|}{\textbf{Macro Average Recall}} &
  \multicolumn{1}{c|}{\textbf{Macro Average F1-score}} &
  \multicolumn{1}{c|}{\textbf{Accuracy}} &
   &
   \\ \hline
\multirow{5}{*}{Extended} &
  \multirow{5}{*}{Global COVID-19 Dataset} &
  Single Hidden Layer NN &
  \multicolumn{1}{l|}{97\%$\pm$0\%} &
  \multicolumn{1}{l|}{97\%$\pm$0\%} &
  \multicolumn{1}{l|}{97\%$\pm$0\%} &
  97\%$\pm$1\% & 7365
   & 491
   \\ \cline{3-9} 
 &
   &
  3 Hidden Layers NN &
  \multicolumn{1}{l|}{97\%$\pm$0\%} &
  \multicolumn{1}{l|}{97\%$\pm$0\%} &
  \multicolumn{1}{l|}{97\%$\pm$0\%} &
  97\%$\pm$1\% & 7755
   & 517
   \\ \cline{3-9} 
 &
   &
  BiLSTM+3 Hidden Layers NN &
  \multicolumn{1}{l|}{97\%$\pm$1\%} &
  \multicolumn{1}{l|}{97\%$\pm$1\%} &
  \multicolumn{1}{l|}{97\%$\pm$1\%} &
  97\%$\pm$1\% & 5709
   & 518
   \\ \cline{3-9} 
 &
   &
  BiLSTM+CNN &
  \multicolumn{1}{l|}{11\%$\pm$5\%} &
  \multicolumn{1}{l|}{33\%$\pm$4\%} &
  \multicolumn{1}{l|}{17\%$\pm$5\%} &
  33\%$\pm$4\% & 8789
   & 536
   \\ \cline{3-9} 
 &
   &
  \textbf{Proposed TRABSA Model} &
  \multicolumn{1}{l|}{\textbf{98\%$\pm$0\%}} &
  \multicolumn{1}{l|}{\textbf{98\%$\pm$0\%}} &
  \multicolumn{1}{l|}{\textbf{98\%$\pm$0\%}} &
  \textbf{98\%$\pm$1\%} & 8288
   & 518
   \\ \hline
\multirow{5}{*}{Extended} &
  \multirow{5}{*}{USA COVID-19 Dataset} &
  Single Hidden Layer NN &
  \multicolumn{1}{l|}{81\%$\pm$3\%} &
  \multicolumn{1}{l|}{81\%$\pm$3\%} &
  \multicolumn{1}{l|}{81\%$\pm$3\%} &
  83\%$\pm$3\% & 870
   & 58
   \\ \cline{3-9} 
 &
   &
  3 Hidden Layers NN &
  \multicolumn{1}{l|}{85\%$\pm$3\%} &
  \multicolumn{1}{l|}{83\%$\pm$1\%} &
  \multicolumn{1}{l|}{84\%$\pm$2\%} &
  85\%$\pm$3\% & 696
   & 58
   \\ \cline{3-9} 
 &
   &
  BiLSTM+3 Hidden Layers NN &
  \multicolumn{1}{l|}{85\%$\pm$1\%} &
  \multicolumn{1}{l|}{85\%$\pm$1\%} &
  \multicolumn{1}{l|}{85\%$\pm$1\%} &
  86\%$\pm$0\% & 1218
   & 59
   \\ \cline{3-9} 
 &
   &
  BiLSTM+CNN &
  \multicolumn{1}{l|}{17\%$\pm$5\%} &
  \multicolumn{1}{l|}{33\%$\pm$5\%} &
  \multicolumn{1}{l|}{22\%$\pm$4\%} &
  51\%$\pm$1\% & 413
   & 59
   \\ \cline{3-9} 
 &
   &
  \textbf{Proposed TRABSA Model} &
  \multicolumn{1}{l|}{\textbf{87\%$\pm$1\%}} &
  \multicolumn{1}{l|}{\textbf{86\%$\pm$1\%}} &
  \multicolumn{1}{l|}{\textbf{86\%$\pm$1\%}} &
  \textbf{87\%$\pm$1\%} & 1081
   & 47
   \\ \hline
\multirow{5}{*}{External} &
  \multirow{5}{*}{Twitter Dataset} &
  Single Hidden Layer NN &
  \multicolumn{1}{l|}{93\%$\pm$3\%} &
  \multicolumn{1}{l|}{93\%$\pm$3\%} &
  \multicolumn{1}{l|}{93\%$\pm$3\%} &
  93\%$\pm$3\% & 9891
   & 495
   \\ \cline{3-9} 
 &
   &
  3 Hidden Layers NN &
  \multicolumn{1}{l|}{92\%$\pm$1\%} &
  \multicolumn{1}{l|}{92\%$\pm$1\%} &
  \multicolumn{1}{l|}{92\%$\pm$1\%} &
  92\%$\pm$1\% & 4608
   & 288
   \\ \cline{3-9} 
 &
   &
  BiLSTM+3 Hidden Layers NN &
  \multicolumn{1}{l|}{92\%$\pm$3\%} &
  \multicolumn{1}{l|}{92\%$\pm$3\%} &
  \multicolumn{1}{l|}{92\%$\pm$3\%} &
  92\%$\pm$3\% & 8700
   & 291
   \\ \cline{3-9} 
 &
   &
  BiLSTM+CNN &
  \multicolumn{1}{l|}{53\%$\pm$2\%} &
  \multicolumn{1}{l|}{49\%$\pm$4\%} &
  \multicolumn{1}{l|}{46\%$\pm$3\%} &
  49\%$\pm$4\% & 1752
   & 292
   \\ \cline{3-9} 
 &
   &
  \textbf{Proposed TRABSA Model} &
  \multicolumn{1}{l|}{\textbf{97\%$\pm$1\%}} &
  \multicolumn{1}{l|}{\textbf{97\%$\pm$1\%}} &
  \multicolumn{1}{l|}{\textbf{97\%$\pm$1\%}} &
  \textbf{97\%$\pm$1\%} & 6602
   & 287
   \\ \hline
\multirow{5}{*}{External} &
  \multirow{5}{*}{Reddit Dataset} &
  Single Hidden Layer NN &
  \multicolumn{1}{l|}{94\%$\pm$3\%} &
  \multicolumn{1}{l|}{93\%$\pm$3\%} &
  \multicolumn{1}{l|}{94\%$\pm$3\%} &
  94\%$\pm$3\% & 1494
   & 90
   \\ \cline{3-9} 
 &
   &
  3 Hidden Layers NN &
  \multicolumn{1}{l|}{94\%$\pm$1\%} &
  \multicolumn{1}{l|}{94\%$\pm$1\%} &
  \multicolumn{1}{l|}{94\%$\pm$1\%} &
  94\%$\pm$2\% & 2415
   & 119
   \\ \cline{3-9} 
 &
   &
  BiLSTM+3 Hidden Layers NN &
  \multicolumn{1}{l|}{94\%$\pm$1\%} &
  \multicolumn{1}{l|}{94\%$\pm$2\%} &
  \multicolumn{1}{l|}{94\%$\pm$2\%} &
  94\%$\pm$1\% & 2464
   & 101
   \\ \cline{3-9} 
 &
   &
  BiLSTM+CNN &
  \multicolumn{1}{l|}{94\%$\pm$1\%} &
  \multicolumn{1}{l|}{94\%$\pm$0\%} &
  \multicolumn{1}{l|}{94\%$\pm$0\%} &
  94\%$\pm$0\% & 1944
   & 119
   \\ \cline{3-9} 
 &
   &
  \textbf{Proposed TRABSA Model} &
  \multicolumn{1}{l|}{\textbf{94\%$\pm$1\%}} &
  \multicolumn{1}{l|}{\textbf{93\%$\pm$0\%}} &
  \multicolumn{1}{l|}{\textbf{94\%$\pm$0\%}} &
  \textbf{95\%$\pm$1\%} & 2200
   & 94
   \\ \hline
\multirow{5}{*}{External} &
  \multirow{5}{*}{Apple Dataset} &
  Single Hidden Layer NN &
  \multicolumn{1}{l|}{81\%$\pm$1\%} &
  \multicolumn{1}{l|}{82\%$\pm$2\%} &
  \multicolumn{1}{l|}{81\%$\pm$1\%} &
  84\%$\pm$3\% & 96
   & 11
   \\ \cline{3-9} 
 &
   &
  3 Hidden Layers NN &
  \multicolumn{1}{l|}{83\%$\pm$2\%} &
  \multicolumn{1}{l|}{81\%$\pm$3\%} &
  \multicolumn{1}{l|}{82\%$\pm$3\%} &
  85\%$\pm$1\% & 55
   & 10
   \\ \cline{3-9} 
 &
   &
  BiLSTM+3 Hidden Layers NN &
  \multicolumn{1}{l|}{87\%$\pm$2\%} &
  \multicolumn{1}{l|}{85\%$\pm$4\%} &
  \multicolumn{1}{l|}{86\%$\pm$3\%} &
  89\%$\pm$0\% & 140
   & 12
   \\ \cline{3-9} 
 &
   &
  BiLSTM+CNN &
  \multicolumn{1}{l|}{85\%$\pm$1\%} &
  \multicolumn{1}{l|}{83\%$\pm$3\%} &
  \multicolumn{1}{l|}{84\%$\pm$2\%} &
  87\%$\pm$0\% & 130
   & 11
   \\ \cline{3-9} 
 &
   &
  \textbf{Proposed TRABSA Model} &
  \multicolumn{1}{l|}{\textbf{88\%$\pm$1\%}} &
  \multicolumn{1}{l|}{\textbf{86\%$\pm$2\%}} &
  \multicolumn{1}{l|}{\textbf{86\%$\pm$2\%}} &
  \textbf{90\%$\pm$0\%} & 210
   & 12
   \\ \hline
\multirow{5}{*}{External} &
  \multirow{5}{*}{US Airline Dataset} &
  Single Hidden Layer NN &
  \multicolumn{1}{l|}{93\%$\pm$3\%} &
  \multicolumn{1}{l|}{93\%$\pm$3\%} &
  \multicolumn{1}{l|}{93\%$\pm$3\%} &
  94\%$\pm$2\% & 1201
   & 48
   \\ \cline{3-9} 
 &
   &
  3 Hidden Layers NN &
  \multicolumn{1}{l|}{94\%$\pm$2\%} &
  \multicolumn{1}{l|}{93\%$\pm$3\%} &
  \multicolumn{1}{l|}{93\%$\pm$3\%} &
  94\%$\pm$2\% & 1166
   & 53
   \\ \cline{3-9} 
 &
   &
  BiLSTM+3 Hidden Layers NN &
  \multicolumn{1}{l|}{94\%$\pm$1\%} &
  \multicolumn{1}{l|}{93\%$\pm$2\%} &
  \multicolumn{1}{l|}{94\%$\pm$1\%} &
  94\%$\pm$1\% & 1012
   & 46
   \\ \cline{3-9} 
 &
   &
  BiLSTM+CNN &
  \multicolumn{1}{l|}{94\%$\pm$3\%} &
  \multicolumn{1}{l|}{93\%$\pm$4\%} &
  \multicolumn{1}{l|}{93\%$\pm$4\%} &
  94\%$\pm$3\% & 842
   & 40
   \\ \cline{3-9} 
 &
   &
  \textbf{Proposed TRABSA Model} &
  \multicolumn{1}{l|}{\textbf{95\%$\pm$0\%}} &
  \multicolumn{1}{l|}{\textbf{95\%$\pm$0\%}} &
  \multicolumn{1}{l|}{\textbf{95\%$\pm$0\%}} &
  \textbf{96\%$\pm$1\%} & 897
   & 39
   \\ \hline
\end{tabular}%
}
\end{table}

The robustness and generalizability of our TRABSA model are evident through its superior performance compared to a wide range of state-of-the-art models used in multiclass sentiment analysis (SA) on Twitter data. Table \ref{tab:literature} summarizes these comparisons, showcasing how TRABSA consistently outperforms models across multiple datasets. Notably, on the global COVID-19 dataset, TRABSA achieves an exceptional macro average precision, recall, F1-score, and accuracy of 98\%, significantly surpassing models such as Jlifi et al.\cite{jlifi_beyond_2024}, which utilized the Ens-RF-BERT approach and achieved a macro average F1-score of 94.03\% and accuracy of 93.01\%. Similarly, the model by Sazan et al.\cite{sazan_advanced_2024}, which employed RoBERTa+fastText, attained an F1-score of 92.05\% but still falls short when compared to TRABSA’s 95\% F1-score on the US Airline dataset. Furthermore, the CNN-LSTM model proposed by Mohbey et al.\cite{mohbey_cnn-lstm-based_2024} achieved 91.24\% F1-score, showcasing a respectable result, but it is outperformed by TRABSA’s 98\% on the global COVID-19 dataset.

\begin{table}[!ht]
\centering
\footnotesize
\caption{Summary of the proposed models in the state-of-the-art tweet sentiment analysis literature}
\label{tab:literature}
\begin{tabular}[!ht]{| l | p{2cm} | p{2cm} |p{2cm}|p{2cm}|p{2cm}| p{1cm} |}
\hline
\textbf{Study} & \textbf{Model Used} & \textbf{Dataset} & \textbf{Macro Average Precision} & \textbf{Macro Average Recall} & \textbf{Macro Average F1-score} & \textbf{Accuracy}\\
\hline
Qi \& Shabrina (2023)\cite{qi_sentiment_2023} & BoW+SVC & UK COVID-19 Twitter Dataset & 69.66\% & 70.33\% & 69.66\% & 71.00\% \\
\hline
\textbf{Ours} & \textbf{TRABSA} & \textbf{UK COVID-19 Twitter Dataset} & \textbf{94.00\%} & \textbf{93.00\%} & \textbf{94.00\%} & \textbf{94.00\%} \\ 
\hline
dos Santos Neto et al., (2024)\cite{dos_santos_neto_survey_2024} & BERT & TripAdvisor & 87.70\% & 88.20\% & 87.90\% & 88.20\% \\
\hline 
Brum \& Volpe Nunes (2018)\cite{brum_building_2018} & BERT & TweetSentBR & 73.27\% & 72.75\% & 72.96\% & 72.75\% \\
\hline 
De Souza et al. (2018)\cite{villavicencio_development_2018} & MultiFiT-Twitter LM & Twitter NPS & 72.43\% & 72.46\% & 72.43\% & 72.46\% \\
\hline 
Pilar et al. (2023)\cite{pilar_novel_2023} & Neighbor-sentiment & InterTASS & 57.76\% & 51.39\% & 54.39\% & 61.35\% \\
\hline 
Su \& Kabala (2023)\cite{su_public_2023} & GloVe100+LSTM & 500k ChatGPT-related Tweets Jan-Mar 2023 & 81.10\% & 81.10\% & 81.10\% & 81.10\% \\
\hline 
Memiş et al. (2024)\cite{memis_comparative_2024} & Multiclass CNN model with pre-trained word embedding & Turkish Financial Tweets & - & - & - & 72.73\% \\
\hline 
Kp et al. (2024)\cite{kp_tweet_2024} & Ensemble classifier & Twitter API Dataset & 91.29\% & 89.65\% & 87.32\% & 93.42\% \\
\hline 
Mohbey et al. (2024)\cite{mohbey_cnn-lstm-based_2024} & CNN-LSTM & Monkeypox Tweets & 91.24\% & 91.24\% & 91.24\% & 91.24\% \\
\hline
Sazan et al., 2024)\cite{sazan_advanced_2024} & RoBERTa+fastText & US Airline Dataset & 92.08\% & 92.02\% & 92.05\% & 92.02\% \\
\hline
\textbf{Ours} & \textbf{TRABSA} & \textbf{US Airline Dataset} & \textbf{95.00\%} & \textbf{95.00\%} & \textbf{95.00\%} & \textbf{96.00\%} \\ 
\hline
Jlifi et al. (2024)\cite{jlifi_beyond_2024} & Ens-RF-BERT & Hashtag Covid19 Tweets & 94.03\% & 93.05\% & 94.03\% & 93.01\% \\
\hline
Bhardwaj et al. (2024)\cite{bhardwaj_sentiment_2024} & BoW+LR & COV19Tweets Dataset & 82.00\% & 81.80\% & 81.60\% & 81.80\% \\
\hline
\textbf{Ours} & \textbf{TRABSA} & \textbf{Global COVID-19 Dataset} & \textbf{98.00\%} & \textbf{98.00\%} & \textbf{98.00\%} & \textbf{98.00\%} \\ 
\hline

\end{tabular}
\end{table}

What sets TRABSA apart is its consistent performance across different datasets, including both domain-specific (e.g., the US Airline dataset, where it achieved 96\% accuracy) and global datasets. In contrast, other models often exhibit variability in performance depending on the dataset or sentiment categories. This ability to generalize across diverse contexts, such as pandemic-related tweets and US airline sentiment, highlights TRABSA's robustness in handling complex multiclass SA tasks. The models compared in this table span various techniques, from traditional models such as BoW+SVC\cite{qi_sentiment_2023} to more modern DL architectures like CNN-LSTM\cite{mohbey_cnn-lstm-based_2024} and transformer-based approaches such as BERT\cite{dos_santos_neto_survey_2024}, yet none achieve the same level of performance as TRABSA.

\subsection*{Statistical Validation}
To assess the performance of our proposed TRABSA model against the other top-performing models benchmarked in this study, we performed a 10-fold cross-validated two-tailed paired t-test, each with 50 epochs, on key evaluation metrics: accuracy, macro average precision, recall, and F1-score. By top-performing model, we refer to the best combination of word embedding and the model obtained from Table \ref{tab3}. Our null hypothesis ($H_0$) stated that there is no significant difference between the performance of each model and the TRABSA model for the respective metrics. In contrast, the alternative hypothesis ($H_1$) posited that a significant difference does exist. We utilized a significance level of $\alpha = 5\%$, with a Bonferroni correction to account for multiple comparisons. The t-test results (see Table \ref{tab:stat}) revealed that for all metrics—accuracy, precision, recall, and F1-score—the TRABSA model showed statistically significant improvements compared to the other models ($p-values < 0.05$ after adjustment) by rejecting the $H_0$. For example, the accuracy of the TRABSA model was significantly higher than that of BoW+RF, with a t-value of 108.7332 and a p-value of $2.39\times10^{-15}$, suggesting a meaningful enhancement in performance. Similarly, the TRABSA model consistently demonstrated superior results with significant t-statistics and p-values for precision and recall. These findings robustly support the efficacy of the TRABSA model in delivering enhanced performance metrics compared to traditional models.

\begin{table}[!ht]
\centering
\footnotesize
\caption{10-fold cross-validated paired t-tests comparing macro-average precision, recall, F1 scores, and accuracy of the top-performing models against the TRABSA model}
\label{tab:stat}
\begin{tabular}{|p{4.5cm}|>{\raggedright\arraybackslash}p{1.4cm}|>{\raggedright\arraybackslash}p{1.4cm}|p{1.3cm}|p{1.3cm}|p{1cm}|p{1.1cm} |p{1.2cm} |p{1.2cm}|} \hline  
\textbf{Model} & \textbf{t-value (Accuracy)} & \textbf{p-value (Accuracy)} & \textbf{t-value (Precision)} & \textbf{p-value (Precision)} & \textbf{t-value (Recall)} & \textbf{p-value (Recall)} & \textbf{t-value (F1 score)} & \textbf{p-value (F1 score)} \\ \hline 
BoW\_RF & 108.7332 & 2.39E-15 & 59.19424 & 5.65E-13 & 135.5109 & 3.30E-16 & 128.4593 & 5.33E-16 \\ \hline 
BoW\_NB & 161.109 & 6.95E-17 & 95.1292 & 7.95E-15 & 120.9821 & 9.15E-16 & 105.8549 & 3.04E-15 \\ \hline 
BoW\_SVM & 85.36398 & 2.10E-14 & 67.73762 & 1.68E-13 & 66.34988 & 2.03E-13 & 51.12919 & 2.10E-12 \\ \hline 
BoW\_GBM & 54.78734 & 1.13E-12 & 35.68181 & 5.28E-11 & 48.87382 & 3.15E-12 & 52.17979 & 1.75E-12 \\ \hline 
BoW\_LGBM & 61.19279 & 4.19E-13 & 29.05617 & 3.30E-10 & 41.47385 & 1.37E-11 & 49.3594 & 2.88E-12 \\ \hline 
BoW\_XGBoost & 91.83215 & 1.09E-14 & 62.80512 & 3.32E-13 & 101.6696 & 4.37E-15 & 96.5374 & 6.96E-15 \\ \hline 
BoW\_Catboost & 82.98394 & 2.71E-14 & 42.12589 & 1.19E-11 & 120.3859 & 9.56E-16 & 63.42425 & 3.04E-13 \\ \hline 
BoW\_LGBM+KNN+MLP & 56.52622 & 8.54E-13 & 39.37306 & 2.19E-11 & 53.2787 & 1.45E-12 & 44.7578 & 6.94E-12 \\ \hline 
BoW\_RF+KNN+MLP & 115.5203 & 1.39E-15 & 54.42813 & 1.20E-12 & 83.80527 & 2.48E-14 & 81.01391 & 3.37E-14 \\ \hline 
BoW\_GBM+RF Stacking Classifier & 95.87232 & 7.41E-15 & 76.63942 & 5.55E-14 & 125.9309 & 6.38E-16 & 104.5914 & 3.39E-15 \\ \hline 
BoW\_GBM+RF Voting Classifier & 126.3773 & 6.18E-16 & 67.23083 & 1.80E-13 & 89.36265 & 1.39E-14 & 74.88783 & 6.83E-14 \\ \hline 
RoBERTa\_Single Hidden Layer NN & 46.94341 & 4.52E-12 & 31.96107 & 1.41E-10 & 41.38057 & 1.40E-11 & 30.30619 & 2.27E-10 \\ \hline 
RoBERTa\_3 Hidden Layers NN & 51.66872 & 1.91E-12 & 33.33187 & 9.70E-11 & 97.90134 & 6.14E-15 & 39.84824 & 1.96E-11 \\ \hline 
RoBERTa\_BiLSTM+3 Hidden Layers NN & 43.79562 & 8.43E-12 & 26.53577 & 7.41E-10 & 56.63986 & 8.39E-13 & 46.28162 & 5.14E-12 \\ \hline 
RoBERTa\_BiLSTM+CNN & 57.76621 & 7.03E-13 & 40.85393 & 1.57E-11 & 45.50036 & 5.98E-12 & 60.10854 & 4.92E-13 \\ \hline

\end{tabular}

\end{table}

\section*{Interpretability Analysis}
\label{xai}
This section discusses the interpretability analysis of the TRABSA model, employing SHAP and LIME techniques to enhance explainability.

\subsection*{SHAP}
A useful tool for deciphering and understanding the results of ML models is the SHapley Additive exPlanations (SHAP) framework \cite{jahin_qamplifynet_2023}. The computation and presentation of the relevance assigned to each characteristic in the prediction process are made easier by utilizing the SHAP Python package. Calculating SHAP values, which measure feature contribution and improve the interpretability of ML models, is essential to the SHAP framework. When the features (\(x\)) are unknown, a SHAP value specifies how to go from the expected or base value \(E[f(x)]\) to the actual output \(f\). Furthermore, by clarifying the direction of the link between features and the target variable, SHAP values shed light on how characteristics affect predictions. A characteristic with a SHAP value of 1 or -1, for example, significantly influences the prediction for a given data point favorably or negatively. On the other hand, a feature that approaches 0 in SHAP value has a negligible contribution to the prediction \cite{jahin_qamplifynet_2023}. A range of graphs are provided by the SHAP framework to help in the understanding of feature contributions and to aid in the interpretation and justification of ML models. The following represents how SHAP values are calculated:
\begin{equation}
    \phi_i(x) = \sum_{S \subseteq N\setminus\{i\}} \frac{|S|!(|N|-|S|-1)!}{|N|!} [f(S \cup \{i\}) - f(S)]
\end{equation} 
Where:\\
\( \phi_i(x) \) = the SHAP value for feature \( i \) in the context of input \( x \),\\
\( N \) = the set of all features,\\
\( S \) = a subset of features excluding \( i \),\\
\( f(S) \) = the model's prediction with features \( S \),\\
\( f(S \cup \{i\}) \) = the model's prediction with features \( S \) and feature \( i \) included.

This formulation encapsulates the incremental contribution of feature \( i \) towards the prediction, considering all possible combinations of features.

\subsubsection*{SHAP Text Plot}
\begin{figure*}[!ht]
    \centering
    \begin{subfigure}{1\linewidth}
        \includegraphics[width=\linewidth]{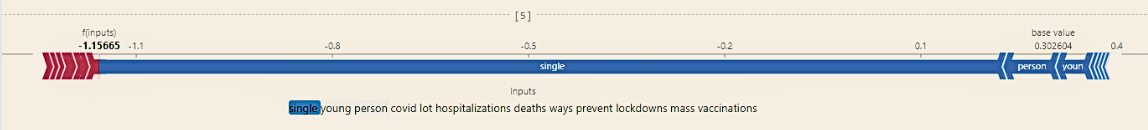}
        \caption{Positive sentiment}
    \end{subfigure}
    \begin{subfigure}{1\linewidth}
        \includegraphics[width=\linewidth]{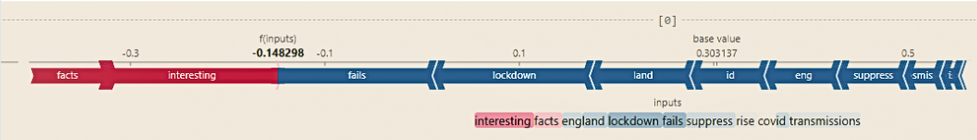}
        \caption{Neutral sentiment}
    \end{subfigure}
    \begin{subfigure}{1\linewidth}
        \includegraphics[width=\linewidth]{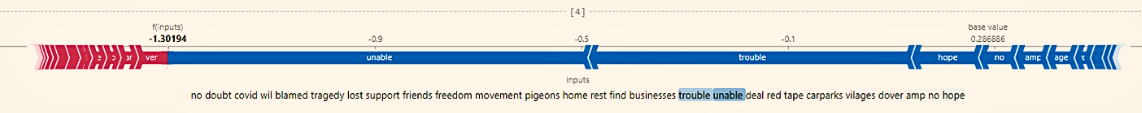}
        \caption{Negative sentiment}
    \end{subfigure}
    \caption{The figure illustrates the importance of each token overlaid on the original text corresponding to that token. It showcases the significance of individual tokens in sentiment prediction, where red regions denote parts of the text increasing the model's output (positive sentiment), while blue regions indicate a decrease in the model's output (negative sentiment).}
    \label{fig:shap1}
\end{figure*}

A thorough illustration of how specific tokens inside a text instance affect a TRABSA model's output may be seen in Figure \ref{fig:shap1}. The graphic illustrates sections that boost (in red) or reduce (in blue) the model's sentiment forecast by superimposing significance values over the original text. This makes it possible to comprehend how certain words or phrases fit into the larger feeling the text is trying to convey nuancedly. Furthermore, the hierarchical structure of significance values preserves the structural linkages between tokens, providing insights into intricate interactions within the text. A comprehensive picture of how the text's combined characteristics affect the model's output is given by the force plot that goes with it; positive features raise the prediction while negative features reduce it. By allowing users to investigate the connections between text segments and how those connections affect the model's predictions, interactive capability further improves interpretability. All things considered, the figure makes it easier to comprehend how the TRABSA model makes decisions and helps with text-based data interpretation and analysis.

\subsubsection*{SHAP Bar Plot}
\begin{figure*}[!ht]
    \centering
    \begin{subfigure}{0.5\textwidth}
        \includegraphics[width=\linewidth]{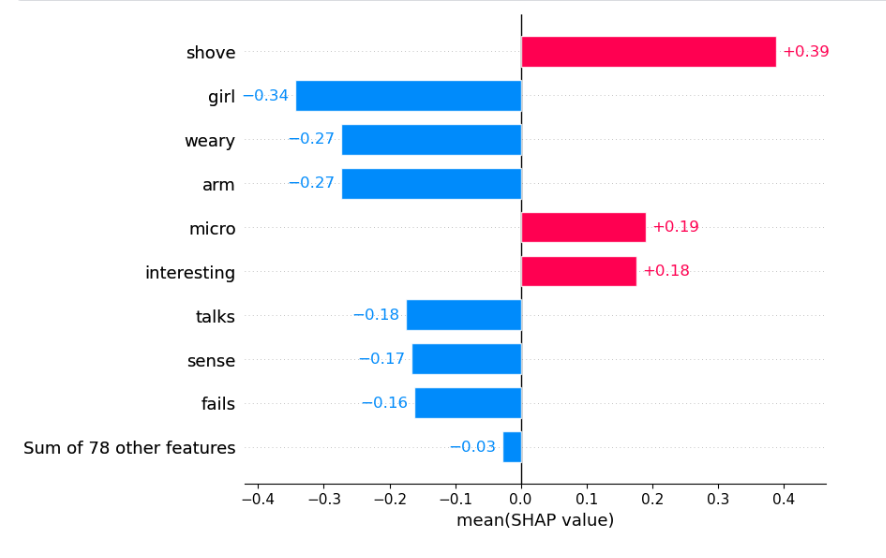}
        \caption{Neutral tokens displayed in their natural order}
    \end{subfigure}
    \begin{subfigure}{0.5\textwidth}
        \includegraphics[width=\linewidth]{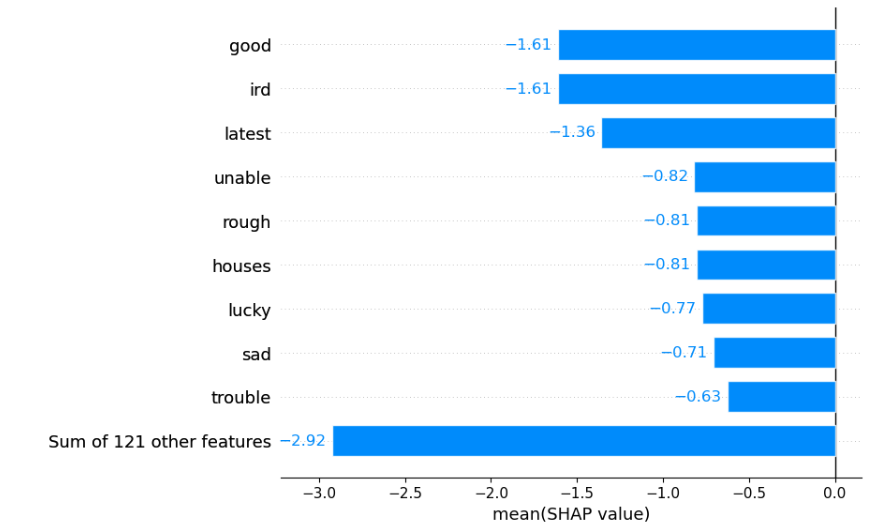}
        \caption{Negative tokens sorted in descending order}
    \end{subfigure}
    \begin{subfigure}{0.5\textwidth}
        \includegraphics[width=\linewidth]{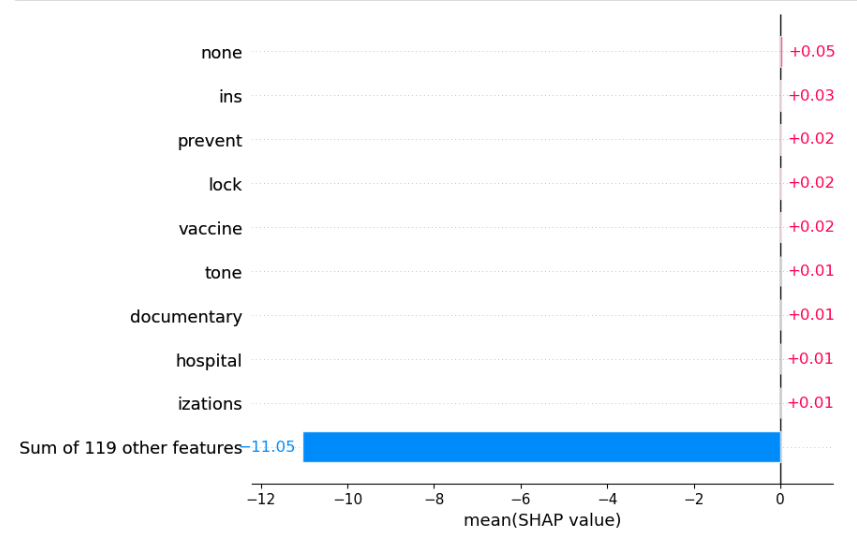}
        \caption{Positive tokens sorted in ascending order}
    \end{subfigure}
    \caption{Summarized importance of tokens in the dataset: (a) Neutral tokens displayed in their natural order, (b) Negative tokens sorted in descending order, and (c) Positive tokens sorted in ascending order. Each bar represents the overall importance of a token, with taller bars indicating greater influence.}
    \label{fig:shap2}
\end{figure*}
The NLP global summaries of the influence of tokens inside a dataset are shown in Figure \ref{fig:shap2}. Every bar in the graphic illustrates how significantly a token affects the model's predictions over the full dataset. Each bar's height represents the influence of the token; higher bars denote greater significance. The plot collects the individual contributions of tokens over numerous instances by compressing the Explanation object over its rows, usually by summing. This method generates a bar chart with as many columns as there were unique tokens in the original dataset by treating each kind of token as a feature. Big groups are split up, and each token gets an equal portion of the total group significance value if the Explanation object contains hierarchical values. Furthermore, arranging the bar chart in a descending sequence helps reveal which tokens have a major impact on the model's predictions.

\subsection*{LIME}
The LIME method provides a methodical technique for evaluating individual predictions given by complicated ML models, which stands for Locally Interpretable Model-Agnostic Explanations \cite{jahin_qamplifynet_2023}. This approach works by estimating the model's behavior around a given forecast. Fundamentally, LIME utilizes a local linear explanation model that follows Equation \ref{eq:lime1} to the letter, making it an additive feature attribution method. LIME presents the idea of "interpretable inputs," which are condensed representations of the original inputs and are represented as \(x_0\). A binary vector of interpretable inputs is mapped to the original input space via the transformation \(x = h_x(x_0)\). Different kinds of mappings \(h_x\) are used for different input spaces. For example, when applied to bag-of-words text characteristics, \(h_x\) translates a binary vector (signaling the presence or absence of words) to the appropriate word count in the source text. LIME aims to minimize the objective function to determine the coefficients \(\phi\):
\begin{equation}
    \xi = \arg\min_{g \in G} \left( L(f, g, \pi_{x_0}) + \Omega(g) \right)
    \label{eq:lime1}
\end{equation}
The loss function in this case is represented by \( L \), which expresses how loyal the explanation model \( g(z_0) \) is to the original model \( f(h_x(z_0)) \). This evaluation is performed across a set of samples weighted by the local kernel \( \pi_{x_0} \) in the reduced input space. Furthermore, the penalty term \( \Omega \) discourages the explanation model \( g \) from being overly complicated. Since \( g \) follows Equation \ref{eq:lime1} and \( L \) uses a squared loss formulation, using penalized linear regression methods is typically necessary to solve Equation \ref{eq:lime1}.

\subsubsection*{LIME Text Explainer Plot}
\begin{figure*}[!ht]
    \centering
    \begin{subfigure}{1\textwidth}
        \includegraphics[width=\linewidth]{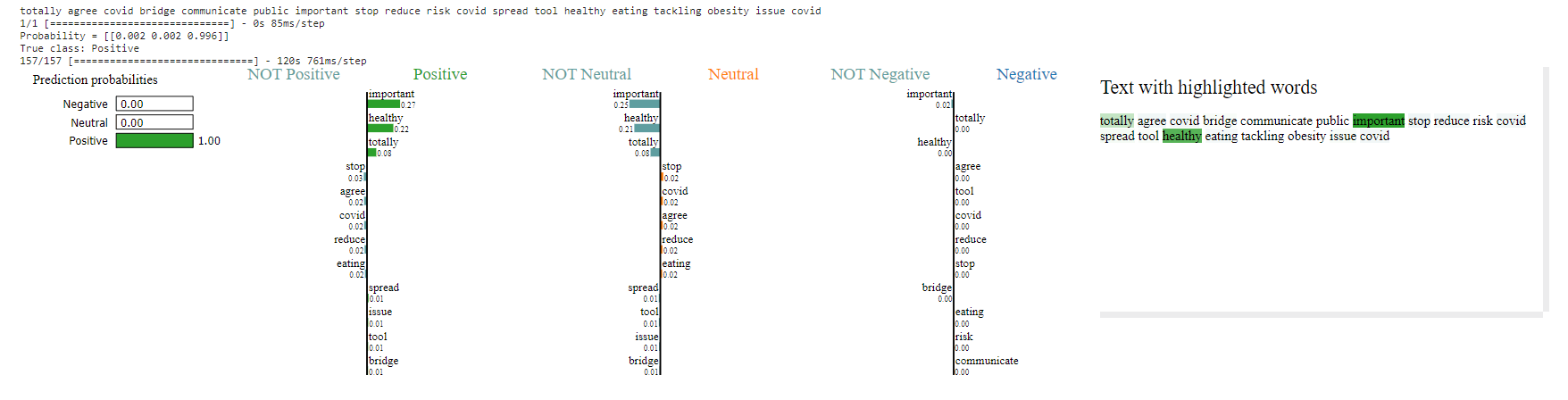}
        \caption{Positive sentiment}
    \end{subfigure}
    \begin{subfigure}{1\textwidth}
        \includegraphics[width=\linewidth]{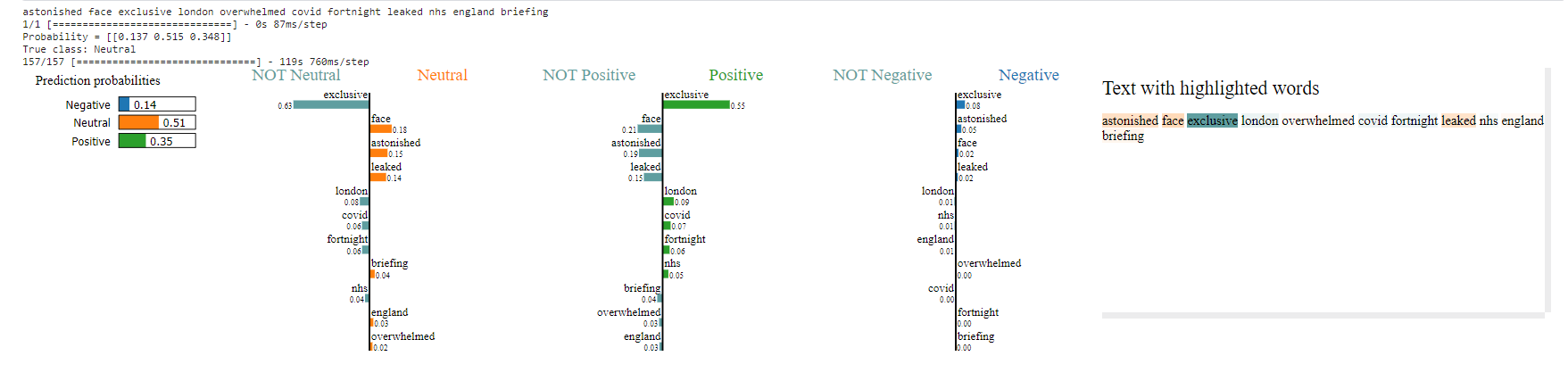}
        \caption{Neutral sentiment}
    \end{subfigure}
    \begin{subfigure}{1\textwidth}
        \includegraphics[width=\linewidth]{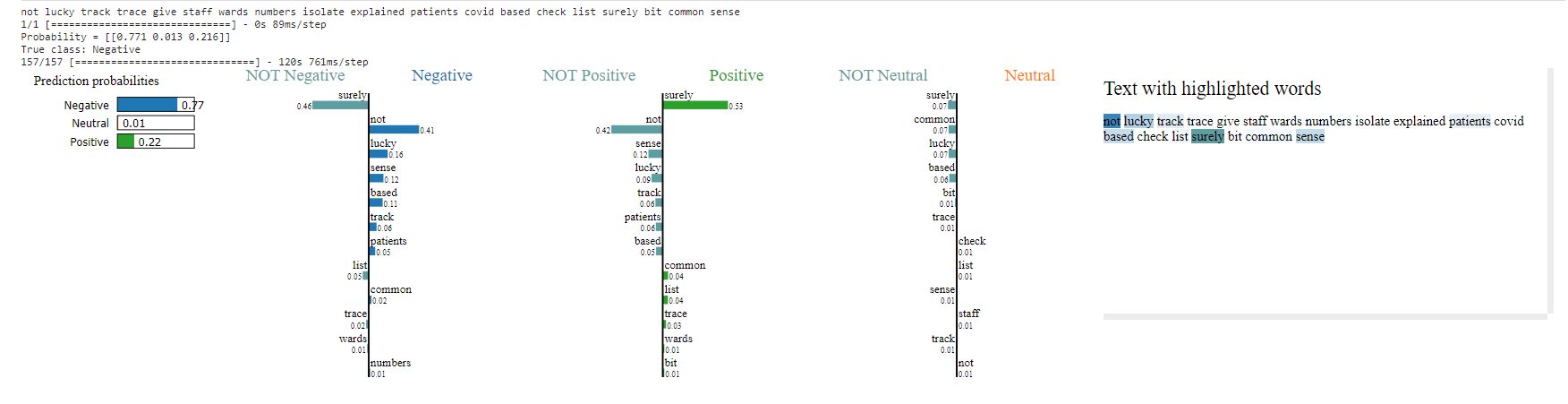}
        \caption{Negative sentiment}
    \end{subfigure}
    \caption{LIME Text Explainer vertical bar plot in descending order of token contributions, illustrating the impact of each token on the TRABSA model's predictions.}
    \label{fig:lime1}
\end{figure*}
The LIME Text Explainer visualization is a useful tool for understanding how particular elements or tokens inside a text affect the model's predictions. Every text token is plotted along the horizontal axis in Figure \ref{fig:lime1}, and its contribution to the prediction is indicated along the vertical axis. Usually, the plot shows how important tokens are by emphasizing how they affect the model's output. Analyzing each token's contribution amount and direction is necessary to understand a LIME Text Explainer plot. Stronger impacts on the model's prediction are indicated by tokens with bigger positive or negative contributions. On the other hand, tokens with contributions closer to zero indicate a negligible influence on the prediction. Additionally, the plot could draw attention to particular words or phrases that greatly impact the model's ability to make decisions. Understanding these influential tokens can provide valuable insights into how the model processes and evaluates textual data.

\section*{Discussions}
\label{discussions}
Our methodology and results demonstrate a significant advancement in SA compared to existing literature. While prior studies have explored diverse models and techniques, our TRABSA model introduces a unique hybrid approach combining transformer-based architectures, attention mechanisms, and BiLSTM networks. This innovative combination enables our model to effectively capture nuanced sentiment patterns, resulting in notably higher accuracy and performance across multiple evaluation metrics than traditional and state-of-the-art transformer models such as BERT and RoBERTa. We conducted a thorough analysis to assess the robustness of our TRABSA model across various datasets and scenarios, consistently observing superior performance across multiple datasets, including extended and external ones. Additionally, our model demonstrates resilience to variations in sentiment expression and context, reaffirming its reliability in diverse real-world scenarios.

This novel hybrid approach offers several benefits to the field, including unparalleled accuracy, robustness, and generalizability across diverse datasets and scenarios. By leveraging the strengths of each component, the TRABSA model can revolutionize SA applications, providing researchers, businesses, and policymakers with deeper insights into public opinion, consumer sentiment, and social trends. Its innovative architecture and superior performance represent a significant advancement in the quest for more accurate and reliable SA tools, with implications extending beyond academic research.

The practical implications of the TRABSA model's advancements are profound, offering tangible benefits across various real-world applications. In market research, the model's ability to accurately analyze sentiment from social media, customer reviews, and other online sources empowers companies to gain valuable insights into consumer preferences, market trends, and brand sentiment. This knowledge informs strategic decision-making processes, product development strategies, and marketing campaigns, ultimately enhancing customer satisfaction and competitive advantage. Furthermore, in social media monitoring and reputation management, the TRABSA model equips organizations with tools to monitor public sentiment, identify emerging issues or crises, and proactively respond to customer feedback in real-time. Detecting and addressing potential issues early on enables businesses to safeguard their reputation and maintain positive relationships with their target audience. Additionally, in the context of public opinion analysis and political discourse, the TRABSA model provides policymakers and analysts with a powerful tool for gauging public sentiment, identifying key concerns, and tracking changes in public perception over time. This knowledge informs policy decisions, communication strategies, and crisis management efforts, ultimately contributing to more informed and responsive governance. The practical applications of the TRABSA model extend across a wide range of industries and domains, offering transformative benefits for businesses, governments, and society as a whole.

\section*{Conclusions and Future Directions}
\label{conclusions}

Our research has yielded significant findings and contributions to SA. We have achieved remarkable results by developing and evaluating the TRABSA model, a novel hybrid approach combining transformer-based architectures, attention mechanisms, and BiLSTM networks. Leveraging the latest RoBERTa-based transformer model and expanding the datasets, we have demonstrated the TRABSA model's exceptional accuracy and relevance, bridging existing gaps in SA benchmarks. Thorough comparisons of word embedding techniques and methodical labeling of tweets using lexicon-based approaches have further enhanced the effectiveness of SA methodologies. Our experiments and benchmarking efforts have highlighted the superiority of the TRABSA model over traditional and state-of-the-art models, showcasing its versatility and robustness across diverse datasets and scenarios. With macro-average precision of 94\%, macro-average recall of 93\%, macro-average F1-score of 94\%, and accuracy of 94\%, our model has proven its efficacy in capturing nuanced sentiment patterns. Additionally, exploring model interpretability techniques using SHAP and LIME has enhanced our understanding and trust in the TRABSA model's predictions, reinforcing its practical applicability.

Despite the significant advancements achieved in our research, several avenues remain for future exploration and improvement in interpretable SA. Firstly, there is scope for refining and expanding model interpretability techniques to provide deeper insights into the factors influencing sentiment predictions. Additionally, integrating multimodal data sources such as text, images, and audio could enhance the richness and accuracy of SA. Addressing ethical considerations regarding bias, fairness, and privacy in SA models is paramount for responsible deployment and usage. Furthermore, exploring the application of SA in emerging domains such as healthcare, finance, and politics could uncover new challenges and opportunities for research and innovation. Overall, continued research in interpretable SA holds the potential to drive meaningful advancements in AI technologies and contribute to more informed decision-making in various fields.

\section*{Data Availability}
\label{data_avail}
The extended datasets, comprising the Global Twitter COVID-19 Dataset and the USA Twitter COVID-19 Dataset, are publicly available for download from the Extended Covid Twitter Datasets (\href{https://data.mendeley.com/datasets/2ynwykrfgf/1}{https://data.mendeley.com/datasets/2ynwykrfgf/1}) repository \cite{jahin_extended_2023}. Additionally, the external datasets used in our research were sourced from Kaggle, including the Twitter and Reddit Dataset (\href{https://www.kaggle.com/datasets/cosmos98/twitter-and-reddit-sentimental-analysis-dataset}{https://www.kaggle.com/datasets/cosmos98/twitter-and-reddit-sentimental-analysis-dataset}), Apple Dataset (\href{https://www.kaggle.com/datasets/seriousran/appletwittersentimenttexts}{https://www.kaggle.com/datasets/seriousran/appletwittersentimenttexts}), and US Airline Dataset \\(\href{https://www.kaggle.com/datasets/crowdflower/twitter-airline-sentiment}{https://www.kaggle.com/datasets/crowdflower/twitter-airline-sentiment}). The code for reproducibility is available in \\
\href{https://github.com/Abrar2652/nlp-roBERTa-biLSTM-attention}{https://github.com/Abrar2652/nlp-roBERTa-biLSTM-attention}.

% \bibliographystyle{nature}
% \bibliography{main}

\section*{Author contributions statement}
M.A.J.: Conceptualization, Methodology, Data curation, Writing - Original Draft Preparation, Software, Visualization, Investigation.
M.S.H.S.: Writing - Original Draft Preparation.
M.F.M.: Conceptualization, Supervision, Reviewing, and Editing. M.R.I.: Conceptualization, Supervision, Reviewing, and Editing. Y.W.: Conceptualization, Supervision, Reviewing.
%J.S.: Funding Acquisition.

\section*{Competing interests}
The authors have no conflict of interest to declare that are relevant to this article.

\section*{Additional information}
Correspondence and requests for materials should be addressed to M.F.M. and M.R.I.

\end{document}